\documentclass[runningheads]{llncs}
\usepackage{cite}
\usepackage{amsmath,amssymb,amsfonts}
\usepackage{algorithmic}
\usepackage{graphicx}
\usepackage{textcomp}
\usepackage{multirow}
\usepackage{subcaption}
\usepackage{framed}
\usepackage[misc,geometry]{ifsym} 
\usepackage{enumitem}
\usepackage[table,xcdraw]{xcolor}
\usepackage{vcell}

\newcommand{\corrAuthor}{$^{\textrm{(\Letter)}}$}
\newcommand{\revpsd}[2]{#2}
\newcommand{\annote}[3]{{%
                \colorbox{#3}{\bfseries\sffamily\footnotesize\textcolor{white}{#2}}
                \color{#3}\footnotesize
                $\blacktriangleright$\textit{#1}$\blacktriangleleft$}
}

\newcommand{\sonja}[1]{\annote{#1}{Sonja}{violet}}

\renewcommand{\annote}[3]{}
\begin{document}
\title{Utility Assessment of Synthetic Data Generation Methods}
%
%
\author{Md Sakib Nizam Khan\inst{1} \corrAuthor 
\and
Niklas Reje\inst{1} 
\and
Sonja Buchegger\inst{1} 
}
\authorrunning{M. Khan et al.}
%
\institute{KTH Royal Institute of Technology, Stockholm, Sweden\\
\email{\{msnkhan \corrAuthor, nreje, buc\}@kth.se}}
\maketitle              
\begin{abstract}

Big data analysis poses the dual problem of privacy preservation and utility, i.e., how accurate data analyses remain after transforming original data in order to protect the privacy of the individuals that the data is about - and whether they are accurate enough to be meaningful. In this paper, we thus investigate across several datasets whether different methods of generating fully synthetic data vary in their utility a priori (when the specific analyses to be performed on the data are not known yet), how closely their results conform to analyses on original data a posteriori, and whether these two effects are correlated. We find some methods (decision-tree based) to perform better than others across the board, sizeable effects of some choices of imputation parameters (notably number of released datasets), no correlation between broad utility metrics  and analysis accuracy, and varying correlations for narrow metrics. We did get promising findings for classification tasks when using synthetic data for training machine learning models, which we consider worth exploring further also in terms of mitigating privacy attacks against ML models such as membership inference and model inversion.

\end{abstract}

\section{Introduction}
In this era of big data and artificial intelligence, technologies are becoming increasingly dependent on data processing and analysis. Since a fair share of the data used by these technologies is the privacy-sensitive data of individuals, there is a growing need for methods that facilitate privacy-preserving data analysis and sharing. Differential Privacy (DP) \cite{dwork2008differential} and k-anonymity-related \cite{sweeney2002k} mechanisms focus on providing \revpsd{privacy guarantees for unlinkability}{provable privacy guarantees}, at the cost of utility \cite{alvim2018local} due to the noise additions required by the mechanisms. \revpsd{}{The mechanisms either transform data/models/results or restrict possible queries from a database to achieve the guarantees. For instance, DP was originally designed for interactive statistical queries to a database and later on extended to non-interactive settings including microdata release. Even though for interactive settings it can provide mathematically quantifiable guarantees, it has been shown that for microdata releases it is not able to provide similar specific confidentiality guarantees \cite{muralidhar2020epsilon}.}  \sonja{reference for utility problem? Reference for and mentioning of interactive setting for DP. I think we should explain that these mechanisms transform data (and/or models/results, restricting queries) to achieve the guarantees.}

\revpsd{A promising alternative is to generate synthetic data from the original data to use for analysis instead.}{Statistical disclosure control (SDC) is another research domain that has been very active in addressing the issue of protecting the confidentiality of individuals while maximizing the utility provided by the released data. An alternative to traditional SDC approaches is publication of synthetic data, an idea that is generally considered as part of SDC literature \cite{ruiz2018privacy}. Synthetic data is generated by drawing from a model that is fitted on original data.} Firstly, with synthetic data, there is no given direct linkability from records to individuals as the records are not \revpsd{}{real \footnote{In \cite{ruiz2018privacy} Ruiz et al. however argue that, based on a scenario where the attacker has access to the original dataset in its entirety,  individuals' representations in synthetic and original datasets remain linked by the information they convey.}}. Secondly, given the large variety of ways to generate synthetic data, there is potential for better utility to arrive at an acceptable level of accuracy for a given analysis.

The main idea of statistical methods for synthetic data generation is imputation \cite{rubin1993statistical} which is originally developed for replacing missing data in survey results. There are different choices of parameters \revpsd{related to imputation}{for the synthesis process} such as which other variables to choose to \revpsd{}{synthesize} one variable, the order of variables chosen for \revpsd{}{synthesis}, etc. which can impact the utility of the generated data. Similarly, depending on the dataset the choice of the synthesizer can have a large impact on the utility of the synthetic data since some synthesizers perform better for numerical variables and some perform better with categorical variables. There has been a lot of research on synthetic data utility in recent years however, \revpsd{there still exists some research gaps}{more research is needed} concerning the impact of different choices during the synthetic data generation process on the utility. Further investigation is also required to find out if there is any correlation between the general utility metrics commonly used to evaluate synthetic data and how well the synthetic data performs for a given analysis.

In this work, we thus investigate the utility provided by different synthetic data generation techniques and synthesis parameters, on separate but similar datasets (Adult and Polish, both census-type data) and on a dataset with different characteristics (Avila). First, we determine the effects on utility as measured by various metrics for the similarity between the original and the synthetic data and, second, by comparing the similarity of results from analyses performed on the original versus on the synthetic data. Third, we investigate to what extent the first can predict the second when the analysis is not known beforehand. In summary, our main contributions are:
\vspace{-5pt}
\begin{itemize}
    \item 
    A comprehensive evaluation of synthetic data utility using three publicly available datasets.
    \item 
    Identification of the individual impact of the choice of variables during synthesis, synthesis order, proper and non-proper synthesis, and number of datasets on the utility of synthetic data.
    \item 
    A comprehensive study on the correlation between different utility metrics and how well the synthetic data performs for any given analysis.
\end{itemize}   
\vspace{-5pt}
\textbf{Organization.} The rest of the paper is organized as follows. In Section \ref{sec:related_work}, we discuss the related work, followed by an overview on the synthetic data generation process and the utility metrics in Section \ref{sec:synthetic_data_generation}. We then present our experimental results in Section \ref{sec:experimental_results}, followed by our discussion and conclusion in Section \ref{sec:dis_conc}.

\section{Related Work}
\label{sec:related_work}
There have been various research works concerning synthetic data generation and the utility provided by them. The related works that we have distilled from the literature mostly focus on the comparison between imputation methods (single and multiple imputation \cite{taub2020impact}, hierarchical Bayes and conventional generalized linear imputation models \cite{graham2009multiply}), evaluation of different mechanisms and tools for synthetic data generation (fully and partially synthetic data\cite{Dandekar2018}, tools for regression \cite{hittmeir2019utility} and classification \cite{hittmeir2019onutility} tasks), evaluation of the usefulness of synthetic data \cite{nowok2015utility, el2021evaluating, wang2019generating, drechsler2008new, lee2013regression}, and what utility metrics to use for comparing original and synthetic data \cite{dankar2022multi, snoke2018general}. In this section, we review some of these works in detail.

Among the works on imputation methods, Taub et al. \cite{taub2020impact} evaluated the real word analyses replicability of singly and multiply imputed CART generated synthetic data using Purdam and Elliot's \cite{purdam2007case} methodology. To evaluate the impact of disclosure control on the analysis outcomes, Purdam and Elliot \cite{purdam2007case} replicated the published analysis on datasets using the disclosure controlled versions of the same datasets. Based on this methodology, Taub et al. \cite{taub2020impact} replicated 9 different sets of analyses involving 28 different tests and models. According to the findings of the authors, multiply imputation performed better for some analyses, but not for all, and depending on the complicacy of the analysis, single imputation can be useful in some scenarios. The authors also investigated the relationship between the utility metrics and how well a synthetic dataset performs for a given analysis and found that there is no clear relationship.

To investigate the usefulness of synthetic data, Nowok \cite{nowok2015utility} evaluated the performance of non-parametric tree-based synthetic data generation methods (Classification and Regression Trees (CART), bagging, and random forests) using synthpop. Besides general utility measures, the authors used some hypothetical analyses to evaluate analysis specific utility of the synthetic data. From the empirical evaluation, the authors conclude that it is possible to produce useful completely synthetic data using automated methods. Similarly, Drechsler et al. \cite{drechsler2008new} replicated an already published analysis of a dataset using synthetic data and found that the regression coefficients of the synthetic and the original were almost identical and concluded that the authors of the published analysis would have drawn the same conclusion using the synthetic dataset. Lee et al. \cite{lee2013regression} used univariate, bivariate, and linear regression-based exploratory data analysis to evaluate the utility of synthetic data generated using CART models and found that for univariate analysis synthetic data results match the original data, whereas for bivariate and linear regression it was not true. According to the authors, the reason for this could be that the CART model underestimates the strong correlation between the variables in the dataset. 

Among the works on utility metrics, Dankar et al. \cite{dankar2022multi}, first classified the available utility metrics for synthetic data comparison into different categories based on the measure they attempt to preserve. Then the authors chose one metric from each category depending on popularity and consistency and used them to compare the utility of four data synthesizers (i.e Data-Synthesizer (DS), Synthetic Data Vault (SDV), Synthpop Parametric, Synthpop Non-parametric). According to the authors, their experimental results show that the Synthpop Non-parametric is the best performing synthesizer overall and provides the best average values across all metrics as well as the best (overall) stability and consistency. Snoke et al. \cite{snoke2018general} investigated how general utility compares to the specific utility for synthetic data and also presented two contrasting example evaluations. The authors conclude that general utility measures can be used to tailor the methods that are used to synthesize datasets for specific purposes whereas specific utility measures can be used for reassurance after performing any standard exploratory analysis on the synthetic data that the data was not misleading.


Even though there have been a lot of studies on different aspects of synthetic data \revpsd{generation, there is still a lack of understanding regarding}{utility, more research is needed to better understand} how choice and order of variables during synthesis, proper and non-proper synthesis, number of datasets, etc. affect the utility of generated synthetic data. Similarly, there is a lack of concrete work on the correlation between commonly used utility metrics for synthetic data and the performance of synthetic data in any given analysis.  

\section{Synthetic Data Generation}
\label{sec:synthetic_data_generation}

\begin{figure*}[htbp]
    \centering
    \includegraphics [width=0.7\linewidth,trim=345 260 150 190,clip]{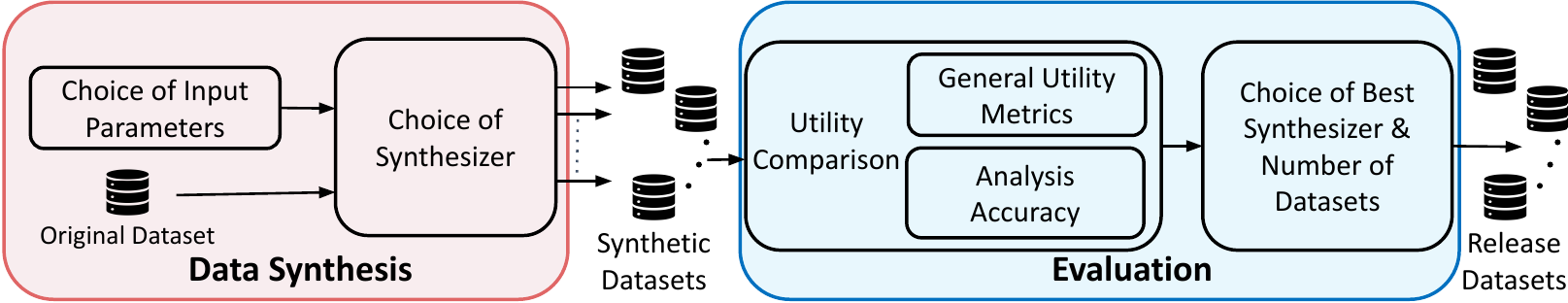}
    \caption{Synthetic Data Generation Process}
    \label{fig:sys_model}
\end{figure*}

Figure \ref{fig:sys_model} provides an overview of synthetic generation process and its two main components, Data Synthesis and Evaluation. 




\vspace{-15pt}

\vspace{-30pt}
\subsection{Data Synthesis}

\subsubsection{Choice of Input Parameters.}
The goal of the data synthesis component is to generate synthetic datasets based on the original dataset. The synthesis process requires a set of input parameters \revpsd{}{alongside the original data}. 

\revpsd{In statistics, imputation is the process of replacing missing data with substitute values drawn from similar records. It was originally developed for solving the problem of missing values in surveys.}{In statistics, imputation is commonly used for solving the problem of missing data in surveys and the idea is to replace them with substitute values drawn from similar records.} Rubin \cite{rubin1987multiple} developed an approach called multiple imputation and later on \cite{rubin1993statistical} proposed to use it for generating fully synthetic datasets. The basic idea proposed by him is to treat the population not selected in a sample as missing data and then use multiple imputation to create synthetic datasets. For multiple imputations, two general approaches have emerged over the years: the joint modeling approach and the fully conditional specification (FCS) approach \cite{Drechsler2011}. With the FCS approach variables are replaced in a specific order whereas with the joint modeling approach multiple variables are replaced together. More details about these approaches can be found in \cite{Drechsler2011}. Here, we present the three main parameter choices for \revpsd{imputation}{synthesis}.




\textbf{Synthesis Order.} For the \revpsd{joint modeling approach, the values in each record are generated at once whereas for}{FCS approach, a particular order of variables needs to be selected for the synthesis.} The order that seems most logical is usually chosen. There are also other approaches \cite{Reiter2005} for choosing the order. Nonetheless, the order needs to be chosen based on the relations between the variables and also the computational complexity. Since the order impacts which relationships are kept between the variables, it can also impact the utility of synthetic data


\textbf{Simple \& Selective Synthesis.} For the FCS approach, it is possible to select only a subset of variables for \revpsd{imputation}{the synthesis} of another variable (we term this approach Selective) instead of using all other variables (which we call Simple).  The Selective approach can be taken in an effort to reduce computation time or to preserve the relationships between variables that are most essential.  However, the downside is that the correlations that are not chosen can get lost, and knowing which relationships to preserve is also not straightforward. 


\textbf{Proper \& Non-proper Synthesis.} \revpsd{In Synthpop, one can chose to use bootstrap samples from the original dataset (Proper) or the entire original dataset (Non-Proper) while generating imputation models}{Depending on whether the new values of the imputation parameters are drawn from their posterior predictive distribution (Proper) or not (Non-proper), synthesis can be of two types.}
\subsubsection{Choice of Synthesizers.}
There are several methods developed over the years for generating synthetic data using multiple imputation. We use three of the most widely used methods: Parametric and Decision Trees based on the FCS approach and Saturated model based on the joint modeling approach.  
The parametric method uses different kinds of regression models depending on what type of data values are meant to be synthesized with linear regression used for numerical variables and logistic regression used for categorical variables. Decision Trees are a non-parametric alternative to the parametric methodology \cite{Nowok2017}.  The Saturated Model approach works by fitting a model that perfectly reproduces the data, to a cross-tabulated table of all the categorical variables \cite{Lee2009, Catall}. Since this method only works with categorical variables, a parametric method or decision tree is needed for the numerical variables. More details on each of these methods can be found in \cite{Reje1438381}.

\vspace{-10pt}
\subsection{Evaluation}
For evaluating the utility of synthetic datasets a common approach found in the literature is to use different utility metrics. Besides this, some studies have also used the accuracy of analysis to measure how well synthetic data imitate the original data for any given analysis. In this section, we briefly discuss these two comparison approaches.   
\vspace{-15pt}
\subsubsection{General Utility Metrics.}
The existing works on synthetic data use different metrics to estimate the utility provided by such data. However, there is no such metric that can individually capture all the different aspects of utility concerning synthetic data. Hence, a combination of different metrics is needed to compare the utility provided by synthetic data and the original data. There are two categories \cite{Karr2006, drechsler2009disclosure} of utility measures which are commonly termed broad measures and narrow measures of utility. The broad measures mostly compare the utility between the entire distributions of synthetic and original data using some statistical distance measures such as Kullback-Leibler (KL) divergence \cite{Karr2006} score. Narrow measures on the other hand tend to compare specific models (e.g., regression, point estimation, etc.) between the synthetic and original dataset and are more widely used in the literature for the utility comparison of synthetic data. Confidence Interval Overlap (CIO) developed by Karr et al. \cite{Karr2006} is a widely used \cite{Drechsler2008,drechsler2009disclosure, taub2020impact} narrow measure of utility for synthetic data. In this work, we use the both CIO and KL divergence as the general utility metrics. 

\textbf{Confidence Interval Overlap (CIO).}
To measure the utility of synthetic data based on confidence interval overlap (CIO),  we use a combination of mean point and regression fit coefficient estimation. \revpsd{}{We calculate the 95\% Confidence Interval (CI) for the mean points and the regression fit coefficients from both the synthetic and the original data and then use the formula proposed by Karr et al in \cite{Karr2006} to measure the CIO}. We calculate the average CIO for all the coefficients of each regression fit and point estimation. We then take the average over all the tests for each $m$ and $k$ where $m$ is number of datasets to release and $k$ is the number tests performed (i.e., the number of different sets of $m$ datasets). \revpsd{These averages are then further averaged over $k$ to get a summary for a specific synthesizer combination and $m$}{}For the point estimates and regression fits, we need to estimate the variance. There are different equations for variance estimation based on combining rules for imputation proposed by Raghunathan et al. \cite{raghunathan2003multiple}. For more details about the imputation combining rules and variance estimation see \cite{Drechsler2018, taub2020impact}. In this work, we use the equation proposed by Raab et al. \cite{raab2016practical} for the variance estimation. The reason for choosing the equation is that it is not dependent on the number of datasets and thus valid even if there is only one synthetic dataset.

\textbf{Kullback-Leibler Divergence.} Kullback-Leibler Divergence is used to measure how close two probability distributions are, based on information entropy and is a global measure of data utility \cite{Woo2009}. We measure the KL divergence score of each variable individually and then take the average over all the variables for the comparison between synthesizers. 



\textbf{Average Percentage Overlap (APO) above 90\%.} A CIO of 90\% or higher has been regarded as good. With the APO measure, we can have a measure of confidence in the utility beyond the average CIO by calculating the ratio of the CIOs over 90\% to all CIOs.


\vspace{-20pt}
\subsubsection{Analysis Accuracy.}

The problem with broad and narrow measures is that none of them is individually sufficient for measuring the utility of synthetic data. They both have certain weaknesses. For instance, narrow measures perform really well for certain analyses but are unable to provide any insights on the overall utility of the dataset whereas broad measures perform well for many analyses but really well for none \cite{Karr2006}. To overcome these issues, multiple studies use either already published analysis \cite{purdam2007case, taub2020impact} on original data or ad hoc analysis \cite{nowok2015utility, Lee2009} to measure the utility provided by synthetic data. Since this approach uses multiple analyses for the utility comparison, it avoids the possibility of reporting unreasonably high or low utility measures based on a single measure which is the case for broad and narrow measures. In this work, for measuring the utility, we use machine learning-based classification accuracy along with the accuracy of multiple ad hoc analyses which compare the multivariate relationship and univariate distributions between synthetic and original data. 
\section{Experimental Results}
\label{sec:experimental_results}
\subsection{Datasets and Experimental Environment}
In this work, for synthetic data generation, we use the R package synthpop developed by Nowok et al. \cite{Nowok2016}. For the datasets, we use three publicly available ones, {\em i.e.,} Polish quality of life dataset (SD2011)~\cite{czapinski2011social} from synthpop example datasets, Adult dataset \cite{kohavi1996adult}, and Avila dataset~\cite{de2018reliable} from UCI Machine Learning Repository \cite{Dua:2019}. While both the Polish and the Adult datasets are standard census surveys, Avila is a completely different type that is not commonly used in existing works on synthetic data generation. We choose Avila because we want to see how well the methods which are mostly developed for census or similar datasets perform for a completely different dataset. More details about the datasets and the modifications that we performed on them can be found in Appendix.

\vspace{-30pt}
\begin{table}[!htbp]
\caption{Naming Convention}
\label{tab:naming_convention}
\resizebox{\textwidth}{!}{
\begin{tabular}{|l|c|c|l|c|c|l|c|}
\hline
\rowcolor[HTML]{C0C0C0} 
\multicolumn{1}{|c|}{\cellcolor[HTML]{C0C0C0}\textbf{Synthesizer}} &
  \textbf{Symbol} &
  \cellcolor[HTML]{EFEFEF} &
  \multicolumn{1}{c|}{\cellcolor[HTML]{C0C0C0}\textbf{Name}} &
  \textbf{Suffix} &
  \cellcolor[HTML]{EFEFEF} &
  \multicolumn{1}{c|}{\cellcolor[HTML]{C0C0C0}} &
  \cellcolor[HTML]{C0C0C0} \\ \cline{1-2} \cline{4-5}
Standard Parametric (SAP) &
  P &
  \cellcolor[HTML]{EFEFEF} &
  Original Ordering &
  No Suffix &
  \cellcolor[HTML]{EFEFEF} &
  \multicolumn{1}{c|}{\multirow{-2}{*}{\cellcolor[HTML]{C0C0C0}\textbf{Name}}} &
  \multirow{-2}{*}{\cellcolor[HTML]{C0C0C0}\textbf{Suffix}} \\ \cline{1-2} \cline{4-5} \cline{7-8} 
\begin{tabular}[c]{@{}l@{}}Decision Tree / Classification \&\\ Regression Tree (CART)\end{tabular} &
  D &
  \cellcolor[HTML]{EFEFEF} &
  Opposite Ordering &
  O &
  \cellcolor[HTML]{EFEFEF} &
   &
   \\ \cline{1-2} \cline{4-5}
\begin{tabular}[c]{@{}l@{}}Saturated Model (Catall) with \\ Predictive Mean Matching (CAP)\end{tabular} &
  CP &
  \cellcolor[HTML]{EFEFEF} &
  Own Ordering &
  V &
  \cellcolor[HTML]{EFEFEF} &
  \multirow{-2}{*}{Non-Proper} &
  \multirow{-2}{*}{No Suffix} \\ \cline{1-2} \cline{4-5} \cline{7-8} 
\begin{tabular}[c]{@{}l@{}}Saturated Model (Catall) with \\ CART (CAC)\end{tabular} &
  CC &
  \cellcolor[HTML]{EFEFEF} &
  \begin{tabular}[c]{@{}l@{}}Largest Categorical \\ Variables First\end{tabular} &
  H &
  \cellcolor[HTML]{EFEFEF} &
   &
   \\ \cline{1-2} \cline{4-5}
Sample &
  S &
  \multirow{-6}{*}{\cellcolor[HTML]{EFEFEF}\textbf{\begin{tabular}[c]{@{}c@{}}Imputation \\ \vspace{30pt} Order\end{tabular}}} &
  \begin{tabular}[c]{@{}l@{}}Largest Categorical \\ Variables Last\end{tabular} &
  L &
  \multirow{-6}{*}{\cellcolor[HTML]{EFEFEF}\textbf{\begin{tabular}[c]{@{}c@{}}Synthesis \\ \vspace{30pt} Method\end{tabular}}} &
  \multirow{-2}{*}{Proper} &
  \multirow{-2}{*}{T} \\ \hline
\end{tabular}
}
\end{table}
\vspace{-10pt}
For the experiments, we generated multiple synthetic datasets for Standard Parametric (P), Cart (D), Catall (i.e., saturated model) with PMM (CP), and Catall with \revpsd{}{Cart (CC) \footnote{We do not preprocess the data since there is no clear benefit of doing so \cite{dankar2021fake}.}}. For P and D, we generated synthetic datasets with 3 different orders, a total of 6 different datasets. Next, for CP, CC, and the 6 datasets for P and D we generated Proper and Non-Proper combinations which sum up to 16 datasets. Additionally, for baseline comparison, we created a fifth synthesizer, Sample, which just samples values from the original dataset for each variable without attempting to retain any relationships between the variables. Thus, for each of the 3 original datasets, we generate 17 dataset collections. \revpsd{}{In the case of synthesis order, just for the sake of comparison, we generated our own ordering where we chose an order that we thought made the most sense concerning preserving relationships between variables}. Table \ref{tab:naming_convention} shows how we symbolize each method and parameter in this work. We use the naming convention shown in the table for the rest of this paper. Our codes for all the experiments are available on GitHub\footnote{https://github.com/sakib570/synthetic-data-utility}. 



\subsection{General Utility Metrics}
\subsubsection{Confidence Interval Overlap (CIO).}
To evaluate the utility of synthetic datasets using confidence interval overlap (CIO), we perform mean point estimation and regression fitting. The mean point estimations are performed on the numerical variables of each dataset. In the case of regression fitting, we perform 20 regression fits on the Polish dataset and 24 regression fits each on the Adult and Avila datasets. The details of regression fits for each dataset can be found in \cite{Reje1438381}. 

\vspace{-5pt}
\paragraph{\textbf{Mean Point Estimation.}} The mean point estimation results for all three datasets show that at least $m = 3$ or higher is required for all variables to reach APO above $90\%$. In terms of synthesizer, though CP performed well for Adult and Polish, it struggled for Avila whereas D and CC performed well for all three datasets. P struggled for both Adult and Avila and was unable to reach APO above $90\%$ regardless of the number of datasets released. The results for mean point estimation can be found in Table \ref{tab:mpe_type_1} under Appendix.  

\begin{figure*}[!h]
\centering
\begin{subfigure}{.45\textwidth}
  \centering
  \includegraphics[width=1\linewidth]{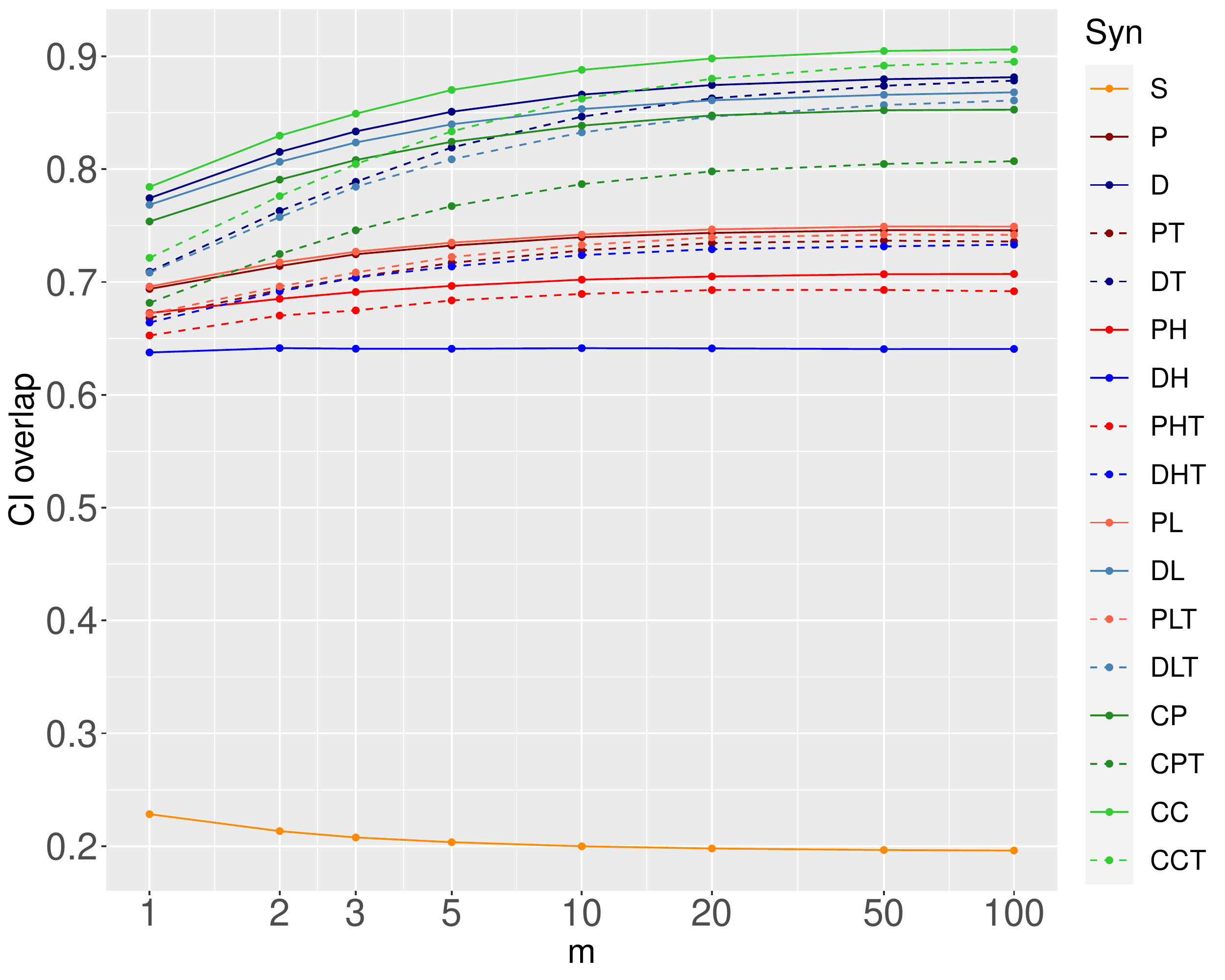}
  \caption{CIO -Polish Dataset}
  \label{fig:polish_rf_t1}
\end{subfigure}\hfil %
\begin{subfigure}{.45\textwidth}
  \centering
  \includegraphics[width=1\linewidth]{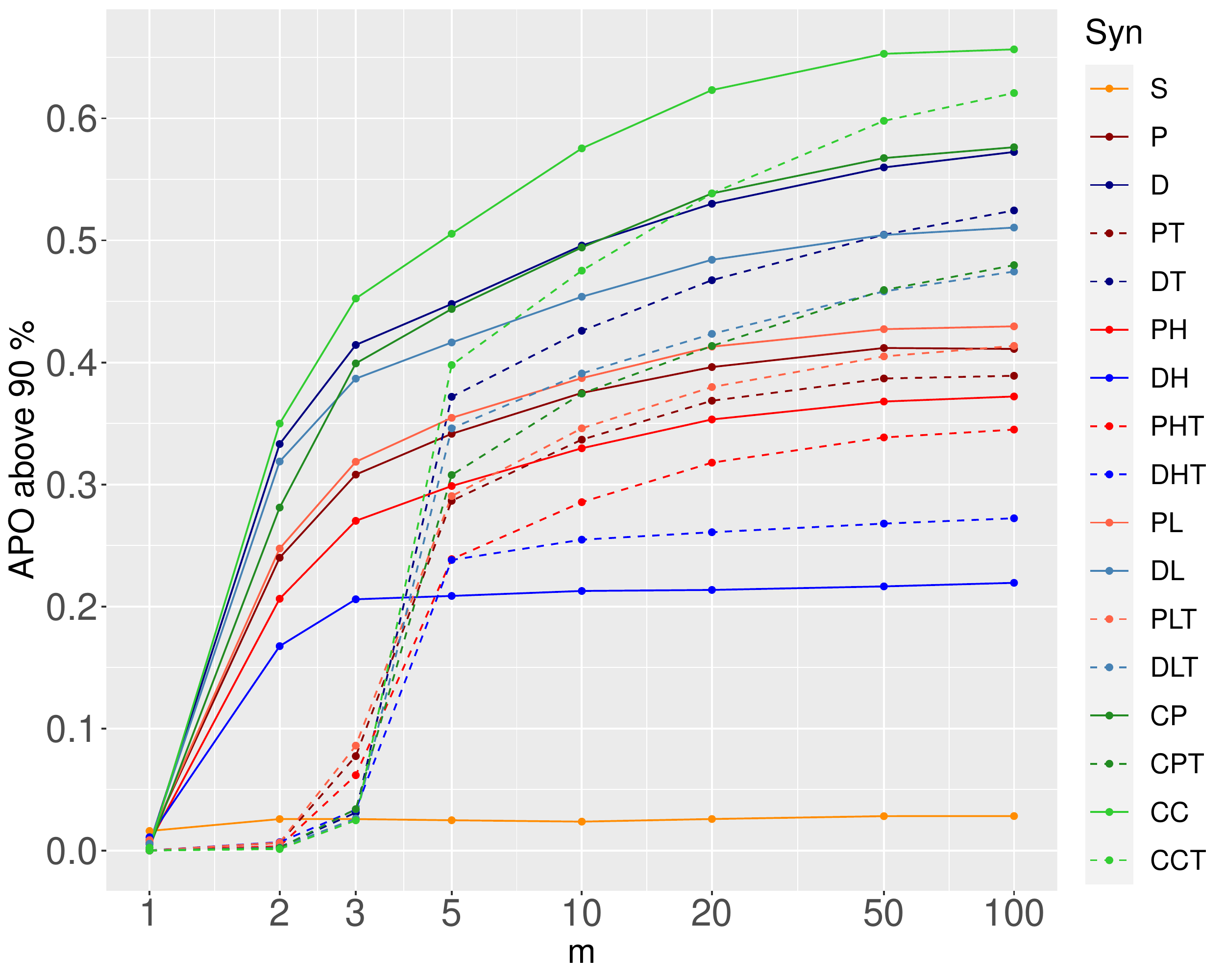}
  \caption{APO above 90\% - Polish Dataset}
  \label{fig:polish_apo_90_t1}
\end{subfigure}\hfil %

\medskip
\begin{subfigure}{.45\textwidth}
  \centering
  \includegraphics[width=1\linewidth]{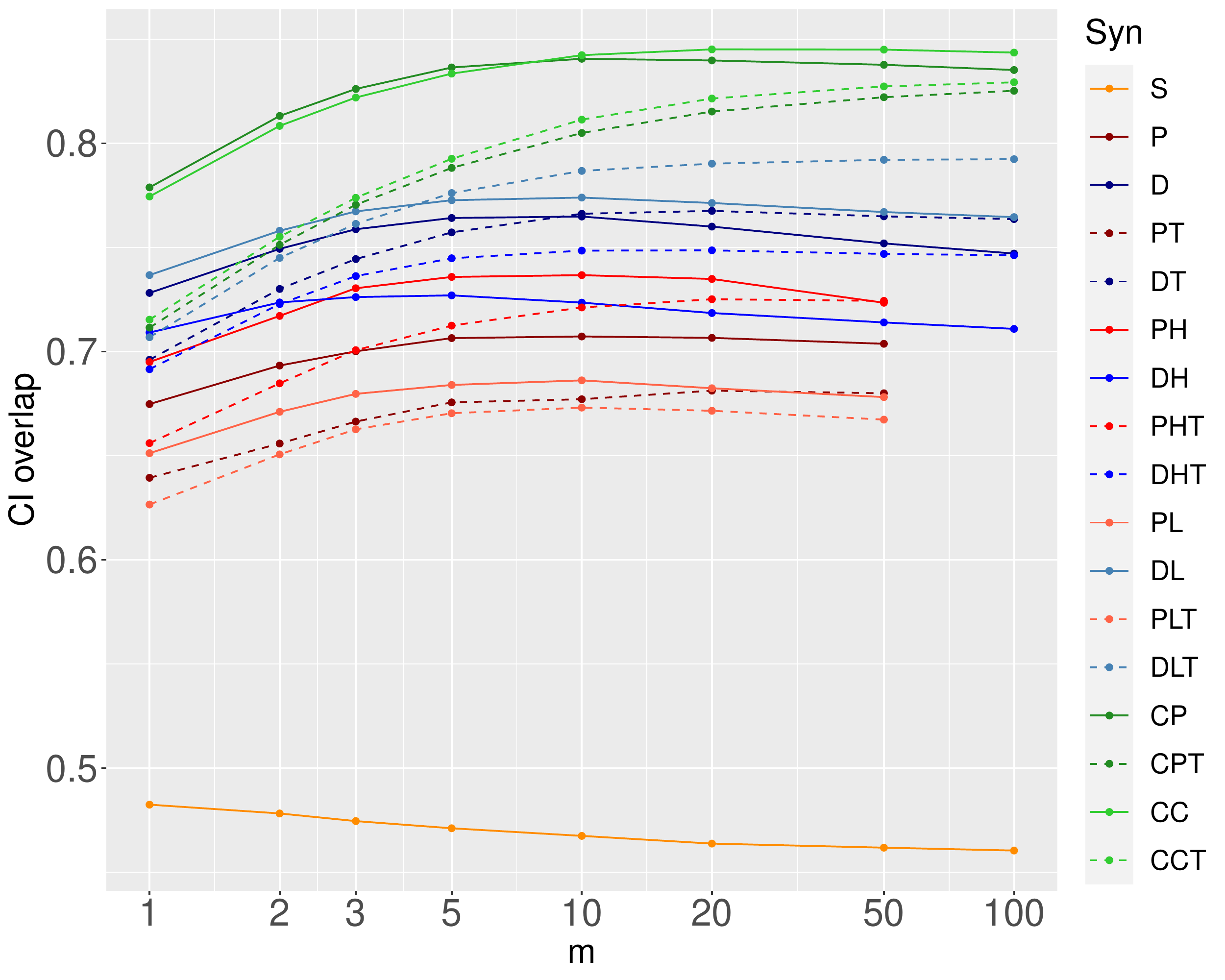}
  \caption{CIO - Adult Dataset}
  \label{fig:adult_rf_t1}
\end{subfigure}\hfil 
\begin{subfigure}{.45\textwidth}
  \centering
  \includegraphics[width=1\linewidth]{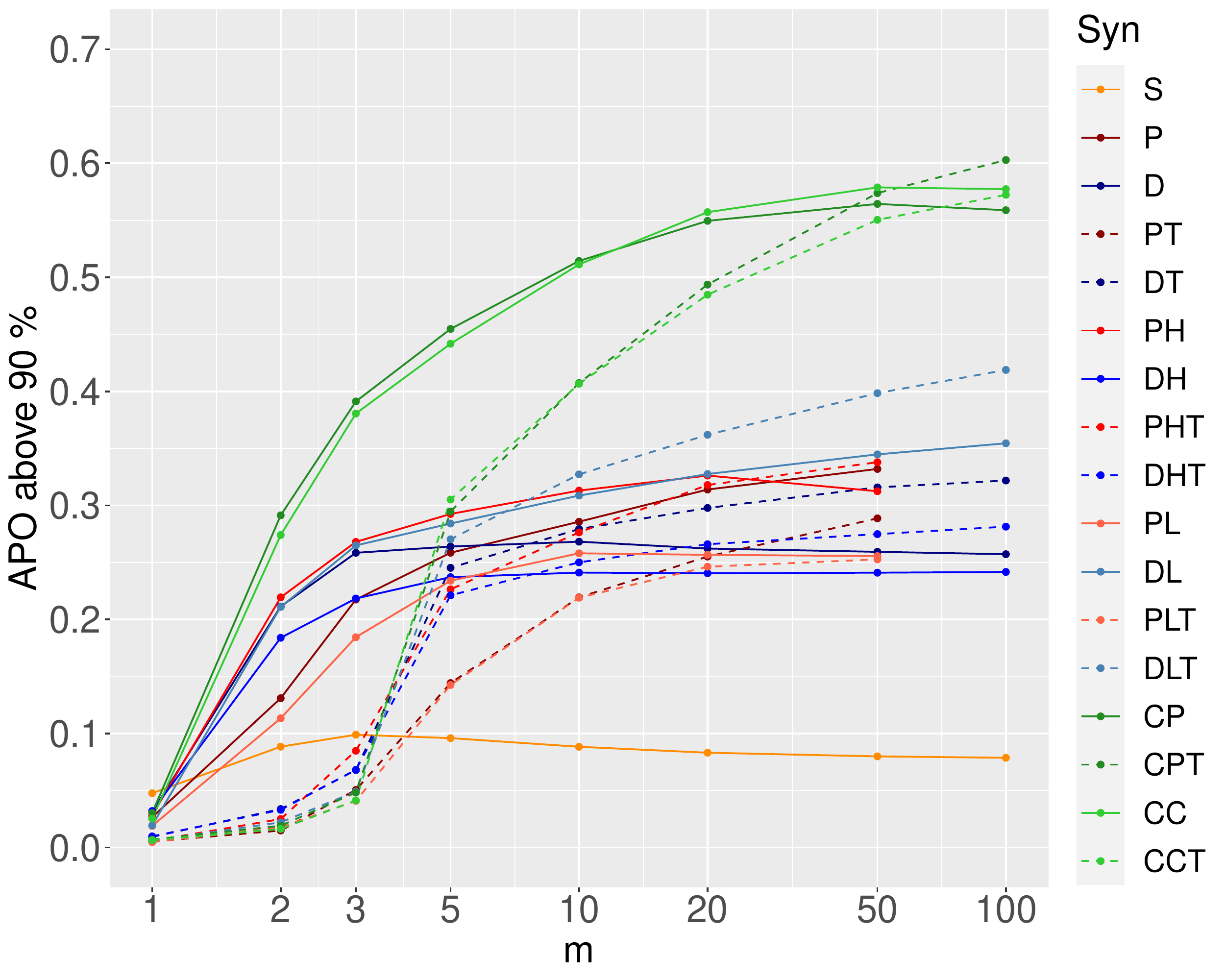}
  \caption{APO above 90\% - Adult Dataset}
  \label{fig:adult_apo_90_t1}
\end{subfigure}\hfil 

\medskip
\begin{subfigure}{.45\textwidth}
  \centering
  \includegraphics[width=1\linewidth]{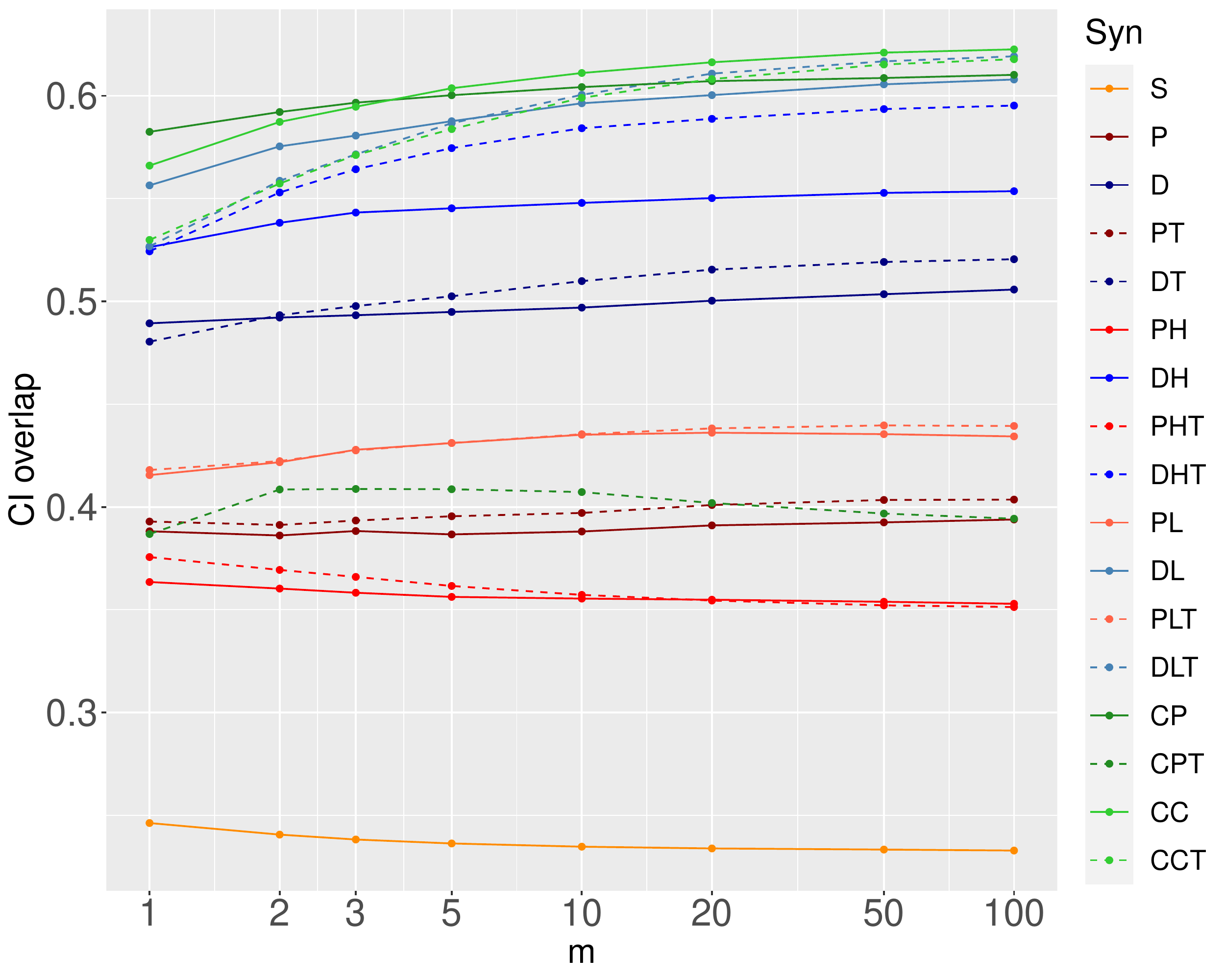}
  \caption{CIO - Avila Dataset}
  \label{fig:avila_rf_t1}
\end{subfigure}\hfil 
\begin{subfigure}{.45\textwidth}
  \centering
  \includegraphics[width=1\linewidth]{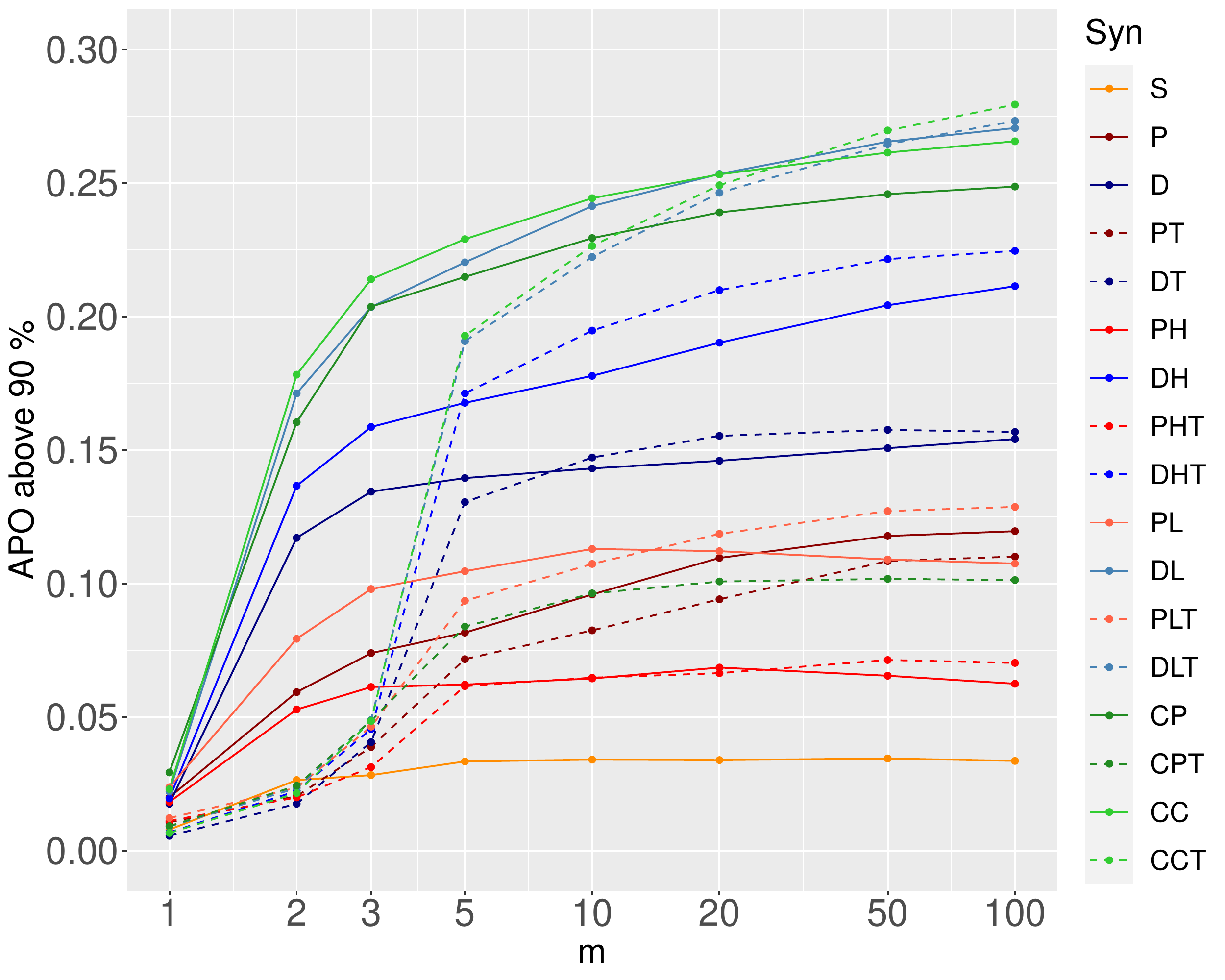}
  \caption{APO above 90\% - Avila Dataset}
  \label{fig:avila_apo_90_t1}
\end{subfigure}\hfil 
\caption{CIO and APO above 90\% for Regression Fits} 
\label{fig:rf_t1}
\end{figure*}

\vspace{-5pt}
\paragraph{\textbf{Regression Fit.}} For regression fit, we calculate both the average CIO and the APO above $90$\% over all the regression fits for each dataset. \revpsd{}{Figure \ref{fig:rf_t1} shows the average CIO (left column) and APO above 90\% (right column) for the regression fits for all three datasets.} 

The average CIO shows that CC is the best synthesizer by being the only one to reach above 90\% for Polish and achieving the highest scores for both Adult and Avila. However, for Adult and Avila, none of the synthesizers can reach above 90\% and in the case of Avila, the best one is only able to reach 60\%. CP and D are the next best ones for all three datasets. The Sample method performs far worse than the others which confirms that the other synthesizers have been able to retain relationships between the variables. We also see that there is a steady increase in overlap as $m$ increases, though there appears to be a somewhat diminishing return with the largest increases being below 10 to 20 datasets and the overlaps above only raising the results a few percentage points at most. A key difference between the results of Polish and Adult/Avila is that for the latter the average CIO drops for many of the synthesizers as $m$ increases, something which only happens for Sample in the case of Polish. Non-proper methods perform better overall than proper ones.  In terms of synthesis order, for Polish the original order (O), for Adult in the case of D the L ordering and for P the H ordering, and for Avila own ordering (V) perform better than other orderings. 


The APO above 90\% also shows similar results as average CIO in terms of synthesizer performance where CC, CP, and D perform well for all three datasets. The major difference between average CIO and APO above 90\% is the results for lower $m$ where APO 90\% shows that lower $m$ (i.e., $m < \sim3$) perform very poorly utility wise compare to higher $m$ which is not the case for average CIO. We can also see that the APO above $90\%$ had a steady increase as the number of datasets to release increases. The increase of APO above $90$\% as $m$ increases is also a lot more than for average CIO and is more significant for higher $m$ as well. For all three datasets for lower $m$ (e.g. $m = \{1,2,3\}$) proper methods perform much worse than Non-Proper methods regardless of synthesizer combinations. In terms of synthesis order, we see similar results as average CIO.

\vspace{-12pt}
\subsubsection{KL-Divergence.}
For the KL-Divergence score (Appendix Table \ref{tab:kl_dv_short}), we calculate the score for all variables and normalize them over the score of Sample and then take the average over all variables for each synthesizer for easier comparison. The KL divergence results conform to the CIO results that CC and D perform well for all three datasets and also CP is not far off. In terms of synthesis order, the original ordering performs better with few exceptions for all three datasets. Similarly, non-proper performs much better than proper in most cases.



\subsection{Analysis Accuracy}
\vspace{-20pt}
\begin{figure}[htbp]
    \centering
    \includegraphics [width=1\linewidth,trim=220 1 220 70,clip]{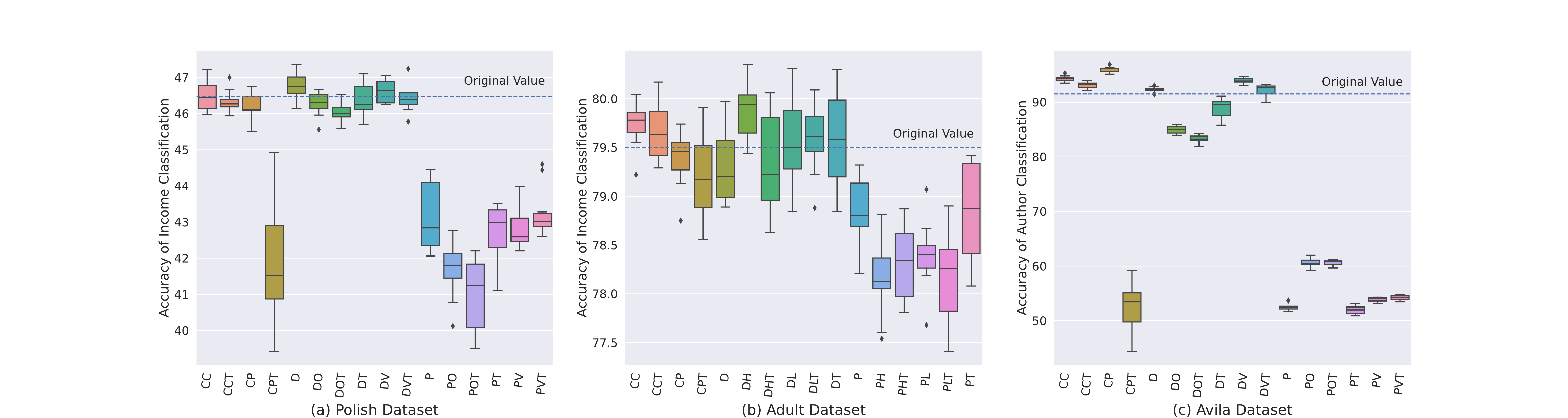}
    \caption{Classification Accuracy}
    \label{fig:classification_accuracy}
\end{figure}
\vspace{-40pt}
\subsubsection{Classification Accuracy.} 
Figure \ref{fig:classification_accuracy} shows the classification accuracy results for all three datasets. From the results we find that CC, CP and D perform well and P is consistently poor for all three datasets. \revpsd{Since, we test the synthetic model accuracy using the original dataset, it is certain that the models trained using the synthetic data can perform well on the real data.}{We tested the synthetic model accuracy using the original dataset and found that the accuracy score is similar to the original model.} We also look at whether the original model and the synthetic model correctly and incorrectly classify same records or they differ. The investigation reveals that for all three datasets the synthetic model and the original model classify the same records correctly and incorrectly and only differ depending on the difference in accuracy score. The details of the classification tasks and model generation process can be found under Appendix.  

\vspace{-15pt}
\begin{figure*}[!htbp]
\centering
\begin{subfigure}{0.33\textwidth}
  \centering
  \includegraphics[width=1\linewidth]{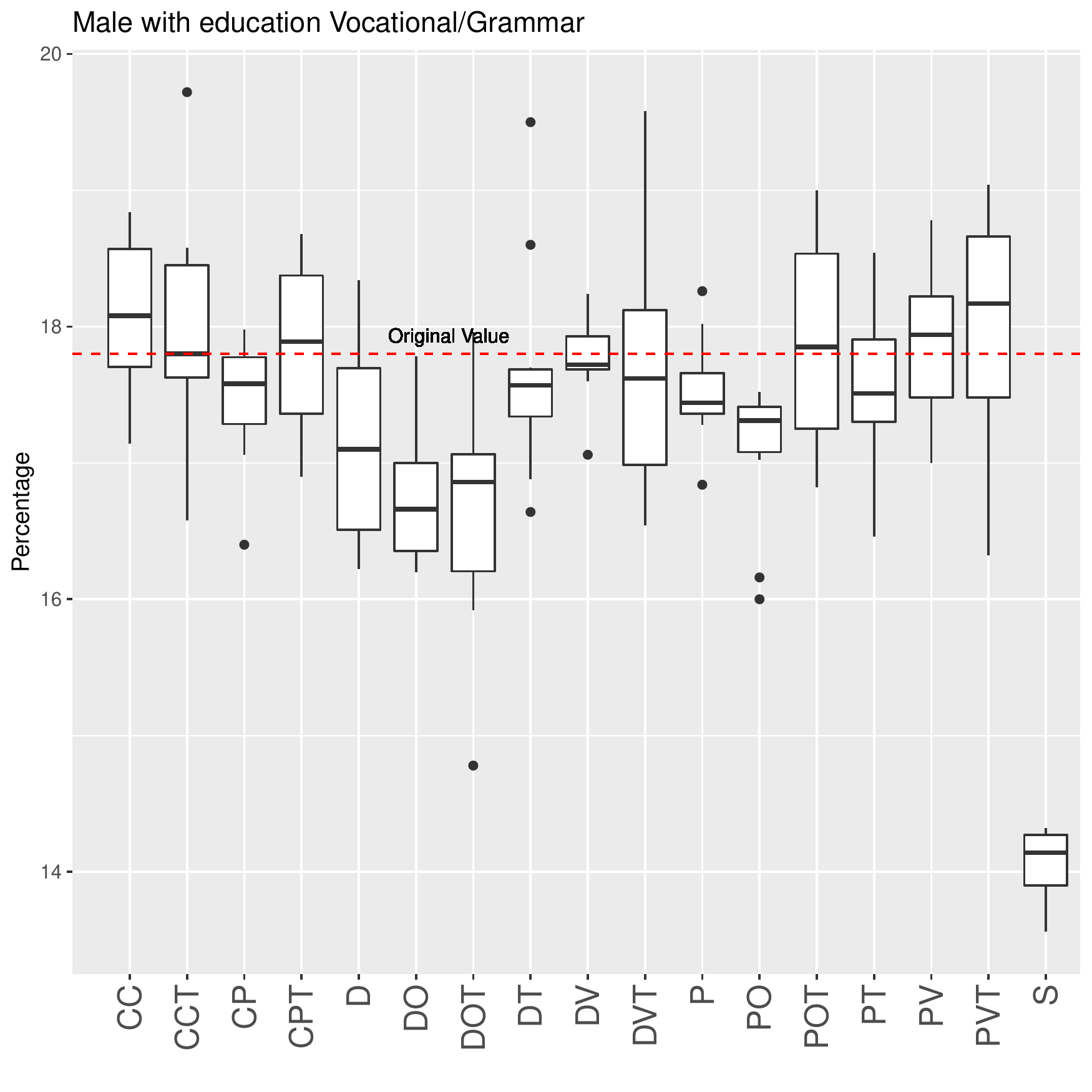}
  \caption{ Polish: Male Education Vocational/Grammar}
  \label{fig:polish_male_edu_voc_gram}
\end{subfigure}%
\begin{subfigure}{.33\textwidth}
  \centering
  \includegraphics[width=1\linewidth]{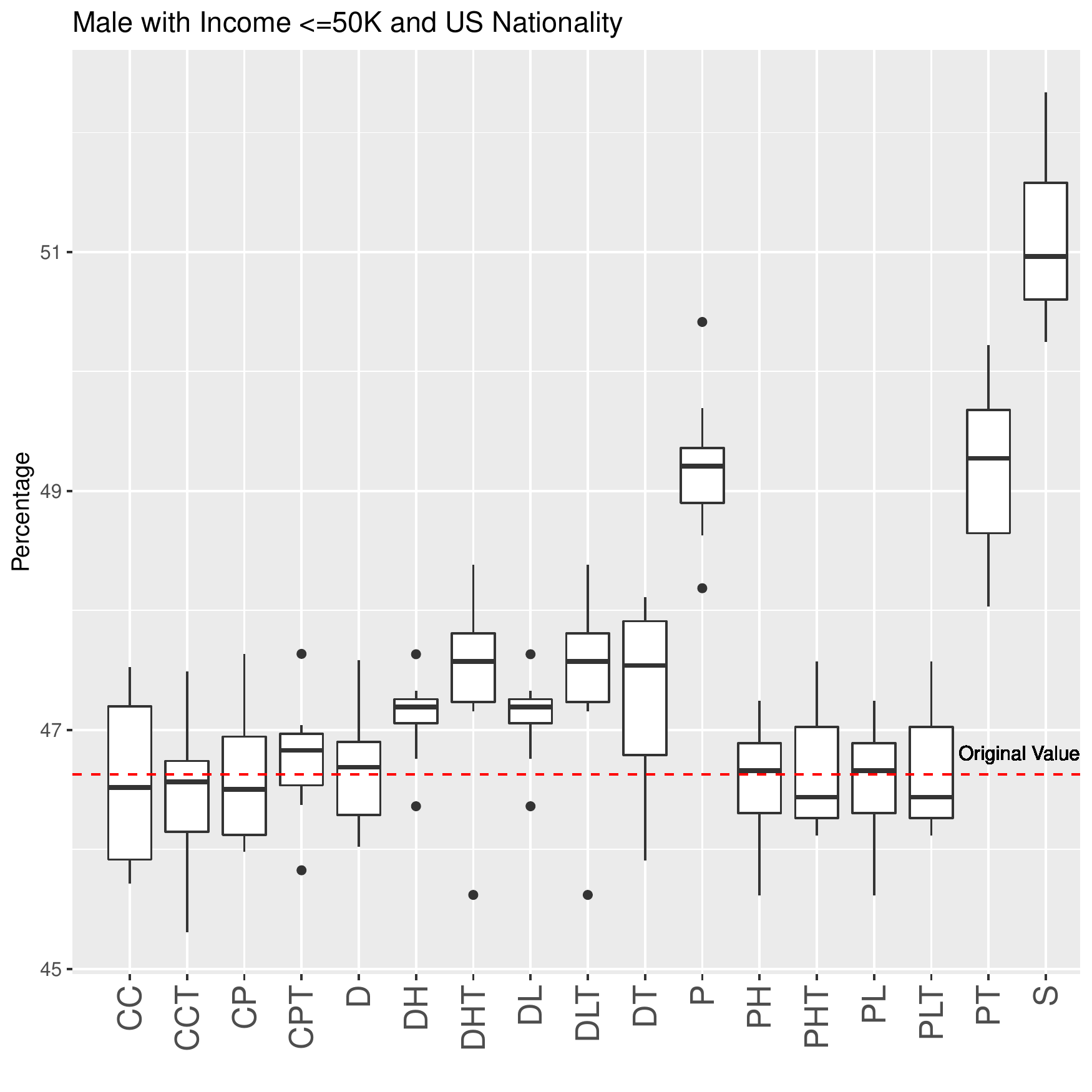}
  \caption{Adult: US Male Income $\leq50K$}
  \label{fig:adult_male_inc_below_50k_us}
\end{subfigure}
\begin{subfigure}{0.33\textwidth}
  \centering
  \includegraphics[width=1\linewidth]{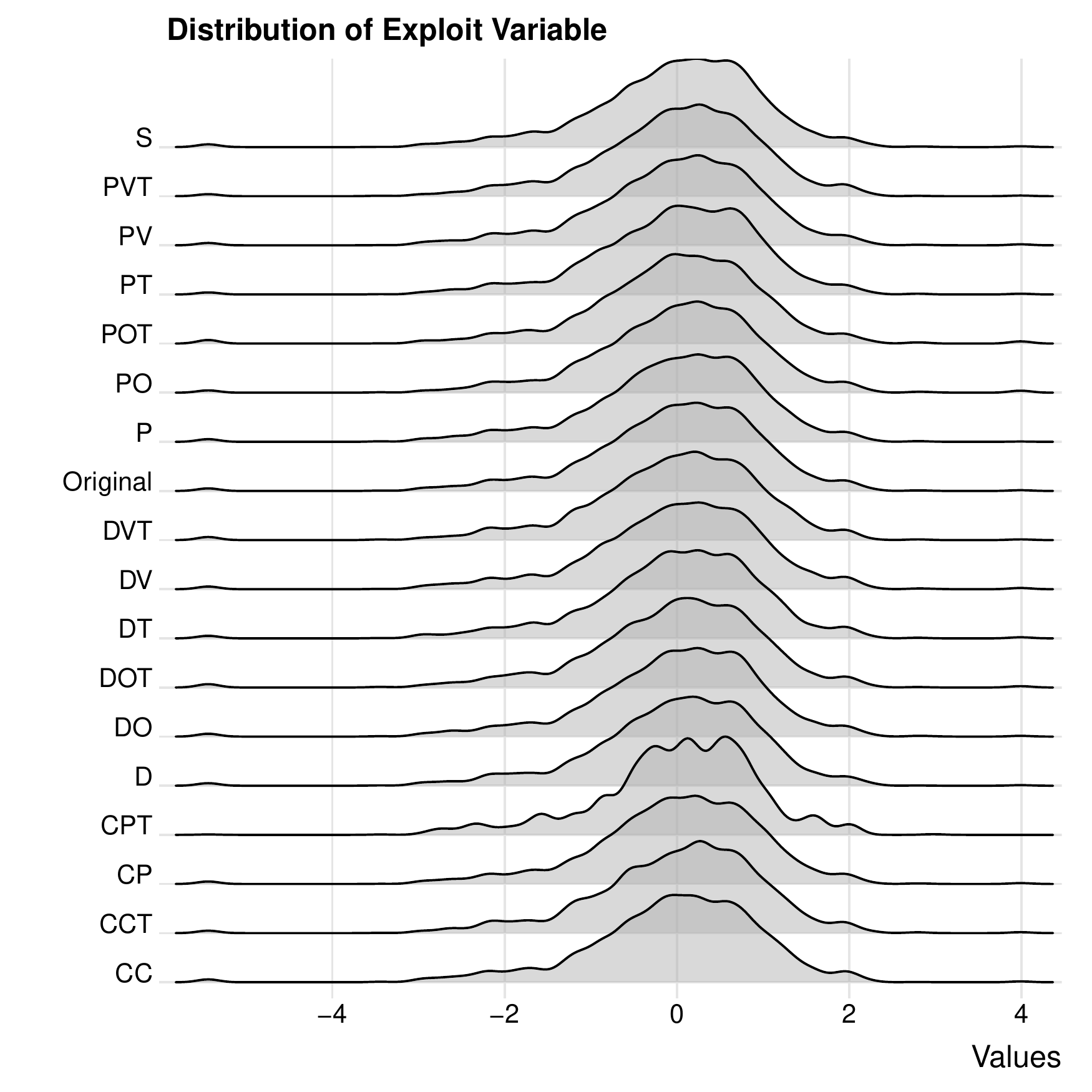}
  \caption{Avila: Exploit Variable Distribution}
  \label{fig:avila_dist_exploit}
\end{subfigure}%
\caption{Ad-hoc Analysis on the Datasets}
\label{fig:adhoc_1}
\end{figure*}
\vspace{-35pt}
\subsubsection{Ad hoc Analysis.}
To determine the effectiveness of synthetic data in different analyses, we look at the multivariate relationship and univariate distributions of the variables present in the datasets. Figure \ref{fig:adhoc_1} shows the results of ad hoc analysis on all three datasets. For Polish, Adult, and Avila, we perform bivariate, multivariate (i.e., 3 variables), and univariate analysis respectively. In the case of Polish and Adult synthesizers CC, CP, and D perform well. In terms of synthesis order, some orders have significant performance improvements over the others for both Polish and Adult. Finally, for Avila, except CPT all synthesizers have similar distributions as the original dataset. In terms of synthesis method, in all three scenarios, we see that non-proper synthesis performs better overall than proper.



\vspace{-25pt}
\begin{table}[!htbp]
\caption{Correlation between Utility Metrics and Analysis Accuracy}
\label{tab:correlation}
\resizebox{\linewidth}{!}{
\begin{tabular}{|llcllcllc|}
\hline
\multicolumn{3}{|c|}{\cellcolor[HTML]{FCEBEA}\textbf{Polish Dataset}} &
  \multicolumn{3}{c|}{\cellcolor[HTML]{ECF4FF}\textbf{Adult Dataset}} &
  \multicolumn{3}{c|}{\cellcolor[HTML]{FFFFE5}\textbf{Avila Dataset}} \\ \hline
\multicolumn{1}{|c|}{\cellcolor[HTML]{FCEBEA}\textbf{Relation}} &
  \multicolumn{1}{c|}{\cellcolor[HTML]{FCEBEA}\textbf{P Value}} &
  \multicolumn{1}{c|}{\cellcolor[HTML]{FCEBEA}\textbf{\begin{tabular}[c]{@{}c@{}}Correlation \\ Co-efficient\end{tabular}}} &
  \multicolumn{1}{c|}{\cellcolor[HTML]{ECF4FF}\textbf{Relation}} &
  \multicolumn{1}{c|}{\cellcolor[HTML]{ECF4FF}\textbf{P Value}} &
  \multicolumn{1}{c|}{\cellcolor[HTML]{ECF4FF}\textbf{\begin{tabular}[c]{@{}c@{}}Correlation \\ Co-efficient\end{tabular}}} &
  \multicolumn{1}{c|}{\cellcolor[HTML]{FFFFE5}\textbf{Relation}} &
  \multicolumn{1}{c|}{\cellcolor[HTML]{FFFFE5}\textbf{P Value}} &
  \cellcolor[HTML]{FFFFE5}\textbf{\begin{tabular}[c]{@{}c@{}}Correlation \\ Co-efficient\end{tabular}} \\ \hline
\rowcolor[HTML]{DFFEDE} 
\multicolumn{9}{|c|}{\cellcolor[HTML]{DFFEDE}\textbf{Correlation between Classification Accuracy and Utility Metrics}} \\ \hline
\rowcolor[HTML]{EFEFEF} 
\multicolumn{1}{|l|}{\cellcolor[HTML]{EFEFEF}\begin{tabular}[c]{@{}l@{}}Accuracy of Income  Classification \\ \& CIO\end{tabular}} &
  \multicolumn{1}{l|}{\cellcolor[HTML]{EFEFEF}1.35E-05} &
  \multicolumn{1}{c|}{\cellcolor[HTML]{EFEFEF}-0.85} &
  \multicolumn{1}{l|}{\cellcolor[HTML]{EFEFEF}\begin{tabular}[c]{@{}l@{}}Accuracy of Income Classification \\ \& CIO\end{tabular}} &
  \multicolumn{1}{l|}{\cellcolor[HTML]{EFEFEF}7.48E-05} &
  \multicolumn{1}{c|}{\cellcolor[HTML]{EFEFEF}-0.81} &
  \multicolumn{1}{l|}{\cellcolor[HTML]{EFEFEF}\begin{tabular}[c]{@{}l@{}}Accuracy of Author Classification \\ \&  CIO\end{tabular}} &
  \multicolumn{1}{l|}{\cellcolor[HTML]{EFEFEF}5.47E-07} &
  -0.91 \\ \hline
\multicolumn{1}{|l|}{\cellcolor[HTML]{FFFFFF}\begin{tabular}[c]{@{}l@{}}Accuracy of Income Classification \\ \& KL Divergence\end{tabular}} &
  \multicolumn{1}{l|}{0.5338} &
  \multicolumn{1}{c|}{-0.16} &
  \multicolumn{1}{l|}{\cellcolor[HTML]{FFFFFF}\begin{tabular}[c]{@{}l@{}}Accuracy of Income Classification \\ \& KL Divergence\end{tabular}} &
  \multicolumn{1}{l|}{0.6213} &
  \multicolumn{1}{c|}{-0.13} &
  \multicolumn{1}{l|}{\cellcolor[HTML]{FFFFFF}\begin{tabular}[c]{@{}l@{}}Accuracy of Author Classification \\ \& KL Divergence\end{tabular}} &
  \multicolumn{1}{l|}{2.43E-02} &
  0.54 \\ \hline
\rowcolor[HTML]{EFEFEF} 
\multicolumn{1}{|l|}{\cellcolor[HTML]{EFEFEF}\begin{tabular}[c]{@{}l@{}}Accuracy of Income Classification \\ \& APO 90\%\end{tabular}} &
  \multicolumn{1}{l|}{\cellcolor[HTML]{EFEFEF}1.54E-03} &
  \multicolumn{1}{c|}{\cellcolor[HTML]{EFEFEF}-0.71} &
  \multicolumn{1}{l|}{\cellcolor[HTML]{EFEFEF}\begin{tabular}[c]{@{}l@{}}Accuracy of Income Classification \\ \& APO 90\%\end{tabular}} &
  \multicolumn{1}{l|}{\cellcolor[HTML]{EFEFEF}2.15E-02} &
  \multicolumn{1}{c|}{\cellcolor[HTML]{EFEFEF}-0.55} &
  \multicolumn{1}{l|}{\cellcolor[HTML]{EFEFEF}\begin{tabular}[c]{@{}l@{}}Accuracy of Author Classification \\ \&  APO 90\%\end{tabular}} &
  \multicolumn{1}{l|}{\cellcolor[HTML]{EFEFEF}3.54E-06} &
  -0.88 \\ \hline
\rowcolor[HTML]{FEDECF} 
\multicolumn{9}{|c|}{\cellcolor[HTML]{FEDECF}\textbf{Correlation between Different Utility Metrics}} \\ \hline
\multicolumn{1}{|l|}{CIO \& APO above 90\%} &
  \multicolumn{1}{l|}{4.14E-07} &
  \multicolumn{1}{c|}{0.91} &
  \multicolumn{1}{l|}{CIO  \&  APO 90\%} &
  \multicolumn{1}{l|}{5.54E-06} &
  \multicolumn{1}{c|}{0.87} &
  \multicolumn{1}{l|}{CIO  \&  APO 90\%} &
  \multicolumn{1}{l|}{3.04E-11} &
  0.98 \\ \hline
\rowcolor[HTML]{EFEFEF} 
\multicolumn{1}{|l|}{\cellcolor[HTML]{EFEFEF}CIO  \& KL  Divergence} &
  \multicolumn{1}{l|}{\cellcolor[HTML]{EFEFEF}3.97E-01} &
  \multicolumn{1}{c|}{\cellcolor[HTML]{EFEFEF}0.22} &
  \multicolumn{1}{l|}{\cellcolor[HTML]{EFEFEF}CIO  \& KL Divergence} &
  \multicolumn{1}{l|}{\cellcolor[HTML]{EFEFEF}1.69E-01} &
  \multicolumn{1}{c|}{\cellcolor[HTML]{EFEFEF}-0.35} &
  \multicolumn{1}{l|}{\cellcolor[HTML]{EFEFEF}CIO  \& KL Divergence} &
  \multicolumn{1}{l|}{\cellcolor[HTML]{EFEFEF}9.80E-02} &
  -0.41 \\ \hline
\multicolumn{1}{|l|}{APO 90\% \& KL Divergence} &
  \multicolumn{1}{l|}{8.43E-01} &
  \multicolumn{1}{c|}{-0.05} &
  \multicolumn{1}{l|}{APO 90\% \& KL Divergence} &
  \multicolumn{1}{l|}{2.28E-01} &
  \multicolumn{1}{c|}{-0.31} &
  \multicolumn{1}{l|}{APO 90\% \& KL Divergence} &
  \multicolumn{1}{l|}{4.66E-02} &
  -0.49 \\ \hline
\rowcolor[HTML]{E4E6FA} 
\multicolumn{9}{|c|}{\cellcolor[HTML]{E4E6FA}\textbf{Correlation between Ad hoc Analysis and Utility Metrics}} \\ \hline
\rowcolor[HTML]{EFEFEF} 
\multicolumn{1}{|l|}{\cellcolor[HTML]{EFEFEF}APO 90\% \&  Female Inc 1K  to 2K} &
  \multicolumn{1}{l|}{\cellcolor[HTML]{EFEFEF}0.06245} &
  \multicolumn{1}{c|}{\cellcolor[HTML]{EFEFEF}-0.46} &
  \multicolumn{1}{l|}{\cellcolor[HTML]{EFEFEF}APO  90\% \&  Male \textless{}=50K  US} &
  \multicolumn{1}{l|}{\cellcolor[HTML]{EFEFEF}8.42E-03} &
  \multicolumn{1}{c|}{\cellcolor[HTML]{EFEFEF}-0.62} &
  \multicolumn{1}{l|}{\cellcolor[HTML]{EFEFEF}APO  90\% \& Units Class A} &
  \multicolumn{1}{l|}{\cellcolor[HTML]{EFEFEF}0.2718} &
  -0.28 \\ \hline
\multicolumn{1}{|l|}{APO 90\% \&  Male Edu Voc/Grm} &
  \multicolumn{1}{l|}{4.80E-04} &
  \multicolumn{1}{c|}{-0.75} &
  \multicolumn{1}{l|}{APO 90\% \& Units Income \textless{}=50K} &
  \multicolumn{1}{l|}{0.4505} &
  \multicolumn{1}{c|}{-0.20} &
  \multicolumn{1}{l|}{CIO \& Units Class A} &
  \multicolumn{1}{l|}{0.3071} &
  -0.26 \\ \hline
\rowcolor[HTML]{EFEFEF} 
\multicolumn{1}{|l|}{\cellcolor[HTML]{EFEFEF}APO  90\% \& Units No Income} &
  \multicolumn{1}{l|}{\cellcolor[HTML]{EFEFEF}0.5931} &
  \multicolumn{1}{c|}{\cellcolor[HTML]{EFEFEF}-0.14} &
  \multicolumn{1}{l|}{\cellcolor[HTML]{EFEFEF}APO 90\% \& White below 50K} &
  \multicolumn{1}{l|}{\cellcolor[HTML]{EFEFEF}0.0776} &
  \multicolumn{1}{c|}{\cellcolor[HTML]{EFEFEF}-0.44} &
  \multicolumn{1}{l|}{\cellcolor[HTML]{EFEFEF}} &
  \multicolumn{1}{l|}{\cellcolor[HTML]{EFEFEF}} &
  \multicolumn{1}{l|}{\cellcolor[HTML]{EFEFEF}} \\ \hline
\multicolumn{1}{|l|}{CIO \&  Female Inc 1K to 2K} &
  \multicolumn{1}{l|}{0.04712} &
  \multicolumn{1}{c|}{-0.49} &
  \multicolumn{1}{l|}{CIO  \&  Male \textless{}=50K US} &
  \multicolumn{1}{l|}{1.12E-04} &
  \multicolumn{1}{c|}{-0.80} &
  \multicolumn{1}{l|}{} &
  \multicolumn{1}{l|}{} &
  \multicolumn{1}{l|}{} \\ \hline
\rowcolor[HTML]{EFEFEF} 
\multicolumn{1}{|l|}{\cellcolor[HTML]{EFEFEF}CIO \&  Male Edu Voc/Grm} &
  \multicolumn{1}{l|}{\cellcolor[HTML]{EFEFEF}1.14E-06} &
  \multicolumn{1}{c|}{\cellcolor[HTML]{EFEFEF}-0.90} &
  \multicolumn{1}{l|}{\cellcolor[HTML]{EFEFEF}CIO \& Units Income \textless{}=50K} &
  \multicolumn{1}{l|}{\cellcolor[HTML]{EFEFEF}0.4568} &
  \multicolumn{1}{c|}{\cellcolor[HTML]{EFEFEF}-0.19} &
  \multicolumn{1}{l|}{\cellcolor[HTML]{EFEFEF}} &
  \multicolumn{1}{l|}{\cellcolor[HTML]{EFEFEF}} &
  \multicolumn{1}{l|}{\cellcolor[HTML]{EFEFEF}} \\ \hline
\multicolumn{1}{|l|}{CIO \& Married alcabuse Yes} &
  \multicolumn{1}{l|}{0.8505} &
  \multicolumn{1}{c|}{-0.05} &
  \multicolumn{1}{l|}{CIO \& White above 50K} &
  \multicolumn{1}{l|}{0.08458} &
  \multicolumn{1}{c|}{-0.43} &
  \multicolumn{1}{l|}{} &
  \multicolumn{1}{l|}{} &
  \multicolumn{1}{l|}{} \\ \hline
\rowcolor[HTML]{EFEFEF} 
\multicolumn{1}{|l|}{\cellcolor[HTML]{EFEFEF}CIO \& Units No Income} &
  \multicolumn{1}{l|}{\cellcolor[HTML]{EFEFEF}0.7308} &
  \multicolumn{1}{c|}{\cellcolor[HTML]{EFEFEF}-0.09} &
  \multicolumn{1}{l|}{\cellcolor[HTML]{EFEFEF}CIO \& White below 50K} &
  \multicolumn{1}{l|}{\cellcolor[HTML]{EFEFEF}0.02481} &
  \multicolumn{1}{c|}{\cellcolor[HTML]{EFEFEF}-0.54} &
  \multicolumn{1}{l|}{\cellcolor[HTML]{EFEFEF}} &
  \multicolumn{1}{l|}{\cellcolor[HTML]{EFEFEF}} &
  \multicolumn{1}{l|}{\cellcolor[HTML]{EFEFEF}} \\ \hline
\end{tabular}
}
\end{table}
\vspace{-25pt}
\subsection{Correlation between Utility Metrics and Analysis Accuracy}
To determine whether there is any relationship between the general utility metrics and analysis accuracy, we conduct several pearson correlation tests between them. For the tests, we measure the correlation of different analysis accuracy results with CIO and APO individually and in some cases with KL divergence as well. We also measure the correlation between the utility metrics CIO, APO, and KL divergence. For the classification accuracy and ad hoc analysis, the relationship is tested between the deviation from the original for each score and the corresponding utility metrics. Table \ref{tab:correlation} shows the results of the correlation tests.

For the classification accuracy, we see that there exists a strong correlation between accuracy score and CIO for all three datasets. The correlation coefficient for all of them is above 0.8 in the negative direction. There is also a moderate correlation between classification accuracy and APO above 90\% with the lowest coefficient being Adult with a value of -0.55. However, when we look at the correlation between classification accuracy and KL divergence score, there is no such consistent correlation. 

In the case of ad hoc analysis, we see a mixed batch of results. For univariate distributions and CIO relationship (i.e., Polish: CIO Raab \& Units No Income, Adult:  CIO Raab \& Units Income $\leq50K$, and Avila: CIO Raab \& Units Class A), we see that there is no strong correlation for all three datasets. However, for multivariate relationships where two or more variables are involved, we see mixed results. For instance, Male Edu Voc/Grm and Male $\leq50K$ US for Polish and Adult respectively show strong correlation with CIO with a correlation coefficient above -0.80. We also see moderate (e.g., Polish: CIO Raab \& Female Inc 1K to 2K, Adult: CIO Raab \& Female Inc 1K to 2K) and no correlation (e.g., Polish: CIO Raab \& Married alcabuse Yes) between the multivariate analysis and CIO. Similar tests with APO above 90\% show that APO either has a similar or lower correlation coefficient than CIO.

The correlation test between the general utility metrics shows that there is a strong correlation between CIO and APO above 90\%. Nonetheless, there seems to be no such correlation between CIO or APO above 90\% with KL divergence. From the correlation test, we find that CIO \revpsd{is a better utility metric}{has a better correlation with analysis accuracy} than APO 90\% and KL divergence. However, CIO also has limitations and it is not possible to guarantee that a higher CIO score will ensure higher accuracy in all types of analysis performed on the synthetic data.

\vspace{-15pt}
\subsection{Generation Time}
\vspace{-5pt}
For generation time, we look at APO above $90$\% with the time it took to generate $100$ datasets for each synthesizer combination for both the Simple and the Selective methodology (Appendix Figure \ref{fig:gen_time}). The result of generation time reveals that CC and D are faster overall and achieve better accuracy. However, CP is much slower (i.e., 40 times slower in some cases) than CC and D though they all achieve similar accuracy. For numerical variables, P performs much faster than other synthesizers, however, the accuracy it achieves is much worse than best performing synthesizers. While Selective is faster for most, it is not significantly faster than Simple overall. The choice of synthesis order can have an huge impact on the generation time in some scenarios (e.g., for P in Adult the slowest order takes 10 times more than the fastest). We do not see any huge difference between proper and non-proper in terms of generation time.

\vspace{-12pt}
\section{Discussion \& Conclusion}
\label{sec:dis_conc}
\vspace{-8pt}
In this work, we look at the impact of different parameters and synthesizers on the utility of synthetic data. We also perform a thorough investigation of the correlation between the utility metrics and the analysis accuracy of synthetic data. Our investigation reveals that the choice of synthetic data generation method, the number of datasets to release, and sometimes the synthesis order can impact the utility of synthetic data. We find that Kulback-Leibler divergence, a broad utility metric, does not correlate with analysis accuracy and some narrow metrics such as confidence-interval overlaps (CIO)  show varying correlations for specific univariate and multivariate analyses. Simply put, these metrics compare data similarity but cannot guarantee that for any given analysis the results are similar enough. Nevertheless, the synthesizers with the best CIOs also performed best in terms of analysis accuracy. We found another more promising effect when comparing the results of classification tasks by machine learning models trained on the synthetic versus on the original data, both tested on original data. There, CIO correlates with how similarly the machine learning models perform the classification task, both in terms of accuracy and which records they classified correctly or incorrectly, respectively. Machine learning models have been shown to be susceptible to privacy threats such as membership inference and model inversion attacks. In future work, we will therefore investigate whether our results generalize in terms of utility and to what extent training on synthetic data can mitigate privacy threats to machine learning models.

\vspace{-5pt}
\section*{Acknowledgment}
\vspace{-5pt}
This work was partially supported by the Wallenberg AI, Autonomous Systems \& Software Program (WASP) funded by Knut \& Alice Wallenberg Foundation.
%
%
%
\vspace{-15pt}
\bibliographystyle{splncs04}
\vspace{-8pt}
\bibliography{references}

\begin{thebibliography}{10}
\providecommand{\url}[1]{\texttt{#1}}
\providecommand{\urlprefix}{URL }
\providecommand{\doi}[1]{https://doi.org/#1}

\bibitem{alvim2018local}
Alvim, M., Chatzikokolakis, K., Palamidessi, C., Pazii, A.: {Local Differential
  Privacy on Metric Spaces: Optimizing the Trade-off with Utility}. In: 2018
  IEEE 31st Computer Security Foundations Symposium (CSF). pp. 262--267. IEEE
  (2018)

\bibitem{czapinski2011social}
Czapi{\'n}ski, J., Panek, T.: {Social Diagnosis 2011. Objective and Subjective
  Quality of Life in Poland}. Contemporary Economics  \textbf{5}(3) (2011)

\bibitem{Dandekar2018}
Dandekar, A., Zen, R.A.M., Bressan, S.: {A Comparative Study of Synthetic
  Dataset Generation Techniques}. In: Hartmann, S., Ma, H., Hameurlain, A.,
  Pernul, G., Wagner, R.R. (eds.) Database and Expert Systems Applications. pp.
  387--395. Springer International Publishing, Cham (2018)

\bibitem{dankar2021fake}
Dankar, F.K., Ibrahim, M.: {Fake It till You Make It: Guidelines for Effective
  Synthetic Data Generation}. Applied Sciences  \textbf{11}(5), ~2158 (2021)

\bibitem{dankar2022multi}
Dankar, F.K., Ibrahim, M.K., Ismail, L.: {A Multi-Dimensional Evaluation of
  Synthetic Data Generators}. IEEE Access  (2022)

\bibitem{de2018reliable}
De~Stefano, C., Maniaci, M., Fontanella, F., di~Freca, A.S.: {Reliable Writer
  Identification in Medieval Manuscripts through Page Layout Features: The
  “Avila” Bible Case}. Engineering Applications of Artificial Intelligence
  \textbf{72},  99--110 (2018)

\bibitem{Drechsler2011}
Drechsler, J.: {Background on Multiply Imputed Synthetic Datasets}, pp. 7--11.
  Springer New York, New York, NY (2011). \doi{10.1007/978-1-4614-0326-5\_2},
  \url{https://doi.org/10.1007/978-1-4614-0326-5\_2}

\bibitem{Drechsler2018}
Drechsler, J.: {Some Clarifications Regarding Fully Synthetic Data}. In:
  Domingo-Ferrer, J., Montes, F. (eds.) Privacy in Statistical Databases. pp.
  109--121. Springer International Publishing, Cham (2018)

\bibitem{Drechsler2008}
Drechsler, J., Bender, S., R\"{a}ssler, S.: {Comparing Fully and Partially
  Synthetic Datasets for Statistical Disclosure Control in the German IAB
  Establishment Panel}. Trans. Data Privacy  \textbf{1}(3),  105–130 (dec
  2008)

\bibitem{drechsler2008new}
Drechsler, J., Dundler, A., Bender, S., R{\"a}ssler, S., Zwick, T.: {A New
  Approach for Disclosure Control in the IAB Establishment Panel — Multiple
  Imputation for a Better Data Access}. AStA Advances in Statistical Analysis
  \textbf{92}(4),  439--458 (2008)

\bibitem{drechsler2009disclosure}
Drechsler, J., Reiter, J.: {Disclosure Risk and Data Utility for Partially
  Synthetic Data: An Empirical Study Using the German IAB Establishment
  Survey}. Journal of Official Statistics  \textbf{25}(4), ~589 (2009)

\bibitem{Dua:2019}
Dua, D., Graff, C.: {UCI Machine Learning Repository} (2017),
  \url{http://archive.ics.uci.edu/ml}

\bibitem{dwork2008differential}
Dwork, C.: {Differential Privacy: A Survey of Results}. In: International
  conference on theory and applications of models of computation. pp. 1--19.
  Springer (2008)

\bibitem{el2021evaluating}
El~Emam, K., Mosquera, L., Jonker, E., Sood, H.: {Evaluating the Utility of
  Synthetic COVID-19 Case Data}. JAMIA open  \textbf{4}(1),  ooab012 (2021)

\bibitem{graham2009multiply}
Graham, P., Young, J., Penny, R.: {Multiply Imputed Synthetic Data: Evaluation
  of Hierarchical Bayesian Imputation Models}. Journal of Official Statistics
  \textbf{25}(2), ~245 (2009)

\bibitem{hittmeir2019onutility}
Hittmeir, M., Ekelhart, A., Mayer, R.: {On the Utility of Synthetic Data: An
  Empirical Evaluation on Machine Learning Tasks}. In: Proceedings of the 14th
  International Conference on Availability, Reliability and Security. pp.~1--6
  (2019)

\bibitem{hittmeir2019utility}
Hittmeir, M., Ekelhart, A., Mayer, R.: {Utility and Privacy Assessments of
  Synthetic Data for Regression Tasks}. In: 2019 IEEE International Conference
  on Big Data (Big Data). pp. 5763--5772. IEEE (2019)

\bibitem{Karr2006}
Karr, A.F., Kohnen, C.N., Oganian, A., Reiter, J.P., Sanil, A.P.: {A Framework
  for Evaluating the Utility of Data Altered to Protect Confidentiality}. The
  American Statistician  \textbf{60}(3),  224--232 (2006).
  \doi{10.1198/000313006X124640},
  \url{https://doi.org/10.1198/000313006X124640}

\bibitem{kohavi1996adult}
Kohavi, R., Becker, B.: {Adult Dataset}. UCI machine learning repository
  \textbf{5}, ~2093 (1996)

\bibitem{Lee2009}
Lee, A.: {Generating Synthetic Microdata from Published Marginal Tables and
  Confidentialised Files}. Statistics New Zealand (2009)

\bibitem{lee2013regression}
Lee, J.H., Kim, I.Y., O'Keefe, C.M.: {On Regression-tree-based Synthetic Data
  Methods for Business Data}. Journal of Privacy and Confidentiality
  \textbf{5}(1) (2013)

\bibitem{muralidhar2020epsilon}
Muralidhar, K., Domingo-Ferrer, J., Mart{\'\i}nez, S.: {$\epsilon$
  -Differential Privacy for Microdata Releases Does Not Guarantee
  Confidentiality (Let Alone Utility)}. In: International Conference on Privacy
  in Statistical Databases. pp. 21--31. Springer (2020)

\bibitem{nowok2015utility}
Nowok, B.: {Utility of Synthetic Microdata Generated Using Tree-based Methods}.
  UNECE Statistical Data Confidentiality Work Session  (2015)

\bibitem{Nowok2016}
Nowok, B., Raab, G., Dibben, C.: {synthpop: Bespoke Creation of Synthetic Data
  in R}. Journal of Statistical Software, Articles  \textbf{74}(11),  1--26
  (2016). \doi{10.18637/jss.v074.i11}, \url{https://www.jstatsoft.org/v074/i11}

\bibitem{Nowok2017}
Nowok, B., Raab, G.M., Dibben, C.: {Providing Bespoke Synthetic Data for the UK
  Longitudinal Studies and Other Sensitive Data with the synthpop Package for
  R}. Statistical Journal of the IAOS  \textbf{33}(3),  785--796 (2017)

\bibitem{Catall}
Nowok, B., Raab, G.M., Dibben, C.: {{synthpop}: Catall} (2019),
  \url{https://cran.r-project.org/web/packages/synthpop/synthpop.pdf\#nameddest=syn.catall}

\bibitem{purdam2007case}
Purdam, K., Elliot, M.: {A Case Study of the Impact of Statistical Disclosure
  Control on Data Quality in the Individual UK Samples of Anonymised Records}.
  Environment and Planning A  \textbf{39}(5),  1101--1118 (2007)

\bibitem{raab2016practical}
Raab, G.M., Nowok, B., Dibben, C.: {Practical Data Synthesis for Large
  Samples}. Journal of Privacy and Confidentiality  \textbf{7}(3),  67--97
  (2016)

\bibitem{raghunathan2003multiple}
Raghunathan, T.E., Reiter, J.P., Rubin, D.B.: {Multiple Imputation for
  Statistical Disclosure Limitation}. Journal of official statistics
  \textbf{19}(1), ~1 (2003)

\bibitem{Reiter2005}
Reiter, J.P.: {Using CART to Generate Partially Synthetic Public Use
  Microdata}. Journal of Official Statistics  \textbf{21},  441--462 (2005)

\bibitem{Reje1438381}
Reje, N.: {Synthetic Data Generation for Anonymization}. Master's thesis, KTH,
  School of Electrical Engineering and Computer Science (EECS) (2020)

\bibitem{rubin1987multiple}
Rubin, D.B.: {Multiple Imputation for Survey Nonresponse} (1987)

\bibitem{rubin1993statistical}
Rubin, D.B.: {Statistical Disclosure Limitation}. Journal of official
  Statistics  \textbf{9}(2),  461--468 (1993)

\bibitem{ruiz2018privacy}
Ruiz, N., Muralidhar, K., Domingo-Ferrer, J.: {On the Privacy Guarantees of
  Synthetic Data: A Reassessment from the Maximum-knowledge Attacker
  Perspective}. In: International Conference on Privacy in Statistical
  Databases. pp. 59--74. Springer (2018)

\bibitem{snoke2018general}
Snoke, J., Raab, G.M., Nowok, B., Dibben, C., Slavkovic, A.: {General and
  Specific Utility Measures for Synthetic Data}. Journal of the Royal
  Statistical Society: Series A (Statistics in Society)  \textbf{181}(3),
  663--688 (2018)

\bibitem{sweeney2002k}
Sweeney, L.: {k-anonymity: A Model for Protecting Privacy}. International
  Journal of Uncertainty, Fuzziness and Knowledge-Based Systems
  \textbf{10}(05),  557--570 (2002)

\bibitem{taub2020impact}
Taub, J., Elliot, M., Sakshaug, J.W.: {The Impact of Synthetic Data Generation
  on Data Utility with Application to the 1991 UK Samples of Anonymised
  Records}. Transactions on Data Privacy  \textbf{13}(1),  1--23 (2020)

\bibitem{wang2019generating}
Wang, Z., Myles, P., Tucker, A.: {Generating and Evaluating Synthetic UK
  Primary Care Data: Preserving Data Utility \& Patient Privacy}. In: 2019 IEEE
  32nd International Symposium on Computer-Based Medical Systems (CBMS). pp.
  126--131. IEEE (2019)

\bibitem{Woo2009}
Woo, M.J., Reiter, J., Oganian, A., Karr, A.: {Global Measures of Data Utility
  for Microdata Masked for Disclosure Limitation}. Journal of Privacy and
  Confidentiality  \textbf{1},  111--124 (2009)

\end{thebibliography}
%
\vspace{-20pt}
\appendix
\section{Appendix}
\label{sec:app}

\textbf{Details of the Datasets}
\begin{itemize}
    \item \textbf{Polish Dataset:} The Polish dataset is from synthpop example datasets which is a Polish census on the quality of life in Poland \cite{czapinski2011social}. We use 14 out of the 35 variables to simplify the computation. The dataset contains records with missing values which are replaced with similar values using random sampling. The total number of records is 5000 for the polish dataset.
    \item \textbf{Adult Dataset:} The Adult dataset \cite{kohavi1996adult} is from UCI Repository intended for predicting whether income exceeds 50K/yr based on census data. The dataset contains 15 variables where we removed numerical education level which is a redundant version of the categorical education variable. We also modify the levels of the native country variable from 44 countries to 7 coarser ones, a change that was required for most of the synthesizers to work. To simplify computation, we use the first 10000 records of the dataset. 
    \item \textbf{Avila Dataset:} The Avila dataset \cite{de2018reliable} is also from UCI Repository intended for predicting the copyist based on the patterns of segments of a Latin bible. The dataset has no missing values and all 11 variables and all 10430 records are used.
\end{itemize}
\vspace{-35pt}
\begin{table*}[!htbp]
\caption{APO above $90$\% for Mean Point Estimations}
\label{tab:mpe_type_1}
\begin{subtable}{.3\linewidth}
\centering
\resizebox{0.8\linewidth}{!}{
\begin{tabular}{r|rrrrr}
\hline
m & S & P & D & CP & CC \\ \hline
1 & 0.00 & 0.08 & 0.00 & 0.00 & 0.00 \\
2 & 0.90 & 0.99 & 0.85 & 0.86 & 0.86 \\
3 & 1.00 & 1.00 & 1.00 & 1.00 & 1.00 \\
5 & 1.00 & 1.00 & 1.00 & 1.00 & 1.00 \\
10 & 1.00 & 1.00 & 1.00 & 1.00 & 1.00 \\
20 & 1.00 & 1.00 & 1.00 & 1.00 & 1.00 \\
50 & 1.00 & 1.00 & 1.00 & 1.00 & 1.00 \\
100 & 1.00 & 1.00 & 1.00 & 1.00 & 1.00 \\ \hline
\end{tabular}
}
\caption{Polish Dataset}
\label{tab:polish_mpe_type_1}
\end{subtable}%
\hfill
\begin{subtable}{.3\linewidth}
\centering
\resizebox{0.8\linewidth}{!}{
\begin{tabular}{r|rrrrr}
\hline
\textbf{m} & \textbf{S} & \textbf{P} & \textbf{D} & \textbf{CP} & \textbf{CC} \\ \hline
1 & 0.01 & 0.00 & 0.00 & 0.01 & 0.00 \\
2 & 0.90 & 0.50 & 0.86 & 0.83 & 0.82 \\
3 & 1.00 & 0.50 & 0.99 & 1.00 & 0.98 \\
4 & 1.00 & 0.50 & 1.00 & 1.00 & 1.00 \\
5 & 1.00 & 0.50 & 1.00 & 1.00 & 1.00 \\
6 & 1.00 & 0.50 & 1.00 & 1.00 & 1.00 \\
7 & 1.00 & 0.50 & 1.00 & 1.00 & 1.00 \\
8 & 1.00 & - & 1.00 & 1.00 & 1.00 \\ \hline
\end{tabular}%
}
\caption{Adult Dataset}
\label{tab:adult_mpe_type_1}
\end{subtable}%
\hfill
\begin{subtable}{.3\linewidth}
\centering
\resizebox{0.8\linewidth}{!}{
\begin{tabular}{r|rrrrr}
\hline
\textbf{m} & \textbf{S} & \textbf{P} & \textbf{D} & \textbf{CP} & \textbf{CC} \\ \hline
1 & 0.05 & 0.08 & 0.06 & 0.14 & 0.05 \\
2 & 0.80 & 0.51 & 0.80 & 0.92 & 0.78 \\
3 & 0.90 & 0.68 & 0.91 & 0.93 & 0.90 \\
5 & 0.95 & 0.75 & 0.96 & 0.92 & 0.95 \\
10 & 0.98 & 0.82 & 0.99 & 0.91 & 0.98 \\
20 & 1.00 & 0.86 & 1.00 & 0.90 & 1.00 \\
50 & 1.00 & 0.89 & 1.00 & 0.89 & 1.00 \\
100 & 1.00 & 0.90 & 1.00 & 0.90 & 1.00 \\ \hline
\end{tabular}
}
\caption{Avila Dataset}
\label{tab:avila_mpe_type_1}
\end{subtable}
\end{table*}
\vspace{-60pt}
\begin{table*}[htbp]
\centering
\caption{KL-Divergence Score}
\label{tab:kl_dv_short}
\resizebox{\textwidth}{!}{%
\begin{tabular}{|l|c|c|c|c|c|c|c|c|c|c|c|c|c|c|c|c|c|c|}
\hline
\rowcolor[HTML]{EFEFEF} 
\multicolumn{1}{|c|}{\cellcolor[HTML]{EFEFEF}\textbf{Dataset}} &
  \textbf{metric} &
  \textbf{S} &
  \textbf{P} &
  \textbf{D} &
  \textbf{PT} &
  \textbf{DT} &
  \textbf{PO} &
  \textbf{DO} &
  \textbf{POT} &
  \textbf{DOT} &
  \textbf{PV} &
  \textbf{DV} &
  \textbf{PVT} &
  \textbf{DVT} &
  \textbf{CP} &
  \textbf{CPT} &
  \cellcolor[HTML]{EFEFEF}\textbf{CC} &
  \textbf{CCT} \\ \hline
Polish &
  \multicolumn{1}{r|}{Average} &
  1.00 &
  1.21 &
  1.06 &
  1.96 &
  2.28 &
  1.13 &
  1.29 &
  1.86 &
  2.33 &
  1.23 &
  1.07 &
  1.96 &
  2.29 &
  1.07 &
  2.37 &
  1.03 &
  2.1 \\ \hline
\rowcolor[HTML]{EFEFEF} 
\multicolumn{1}{|c|}{\cellcolor[HTML]{EFEFEF}\textbf{}} &
  \textbf{} &
  \textbf{S} &
  \textbf{P} &
  \textbf{D} &
  \textbf{PT} &
  \textbf{DT} &
  \textbf{PH} &
  \textbf{DH} &
  \textbf{PHT} &
  \textbf{DHT} &
  \textbf{PL} &
  \textbf{DL} &
  \textbf{PLT} &
  \textbf{DLT} &
  \textbf{CP} &
  \textbf{CPT} &
  \textbf{CC} &
  \textbf{CCT} \\ \hline
Adult &
  \multicolumn{1}{r|}{Average} &
  1.00 &
  9.5 &
  1.04 &
  10.17 &
  2.17 &
  4.72 &
  1.05 &
  5.62 &
  2.21 &
  16.92 &
  1.05 &
  17.21 &
  2.12 &
  1.06 &
  2.48 &
  1.00 &
  2.03 \\ \hline
\rowcolor[HTML]{EFEFEF} 
\multicolumn{1}{|c|}{\cellcolor[HTML]{EFEFEF}\textbf{}} &
  \textbf{} &
  \textbf{S} &
  \textbf{P} &
  \textbf{D} &
  \textbf{PT} &
  \textbf{DT} &
  \textbf{PO} &
  \textbf{DO} &
  \textbf{POT} &
  \textbf{DOT} &
  \textbf{PV} &
  \textbf{DV} &
  \textbf{PVT} &
  \textbf{DVT} &
  \textbf{CP} &
  \textbf{CPT} &
  \textbf{CC} &
  \textbf{CCT} \\ \hline
Avila &
  \multicolumn{1}{r|}{Average} &
  1.00 &
  6.96 &
  1.08 &
  7.32 &
  2.31 &
  8.74 &
  1.6 &
  8.96 &
  2.78 &
  5.7 &
  1.08 &
  5.97 &
  2.28 &
  1.13 &
  25.36 &
  1.08 &
  2.31 \\ \hline
\end{tabular}%
}
\end{table*}
\vspace{-10pt}
\noindent\textbf{Details of the Classification Tasks} Adult and Avila dataset are primarily used for classification tasks using machine learning. The classification task for Adult is to classify whether the income is above or below 50K. Similarly, for Avila it is to classify the author based on the patterns. For Polish dataset we did not find any such common analysis. Since it is a census dataset similar to Adult and has a income variable, we perform a similar income classification on the dataset. For each original dataset, we train the 10 machine learning models using 10 different synthetic datasets and take the average accuracy score. We test the accuracy of each synthetic model using the original dataset.

\begin{figure*}[htbp]
\centering
\resizebox{\linewidth}{!}{
\begin{subfigure}{.33\textwidth}
  \centering
  \includegraphics[width=1\linewidth]{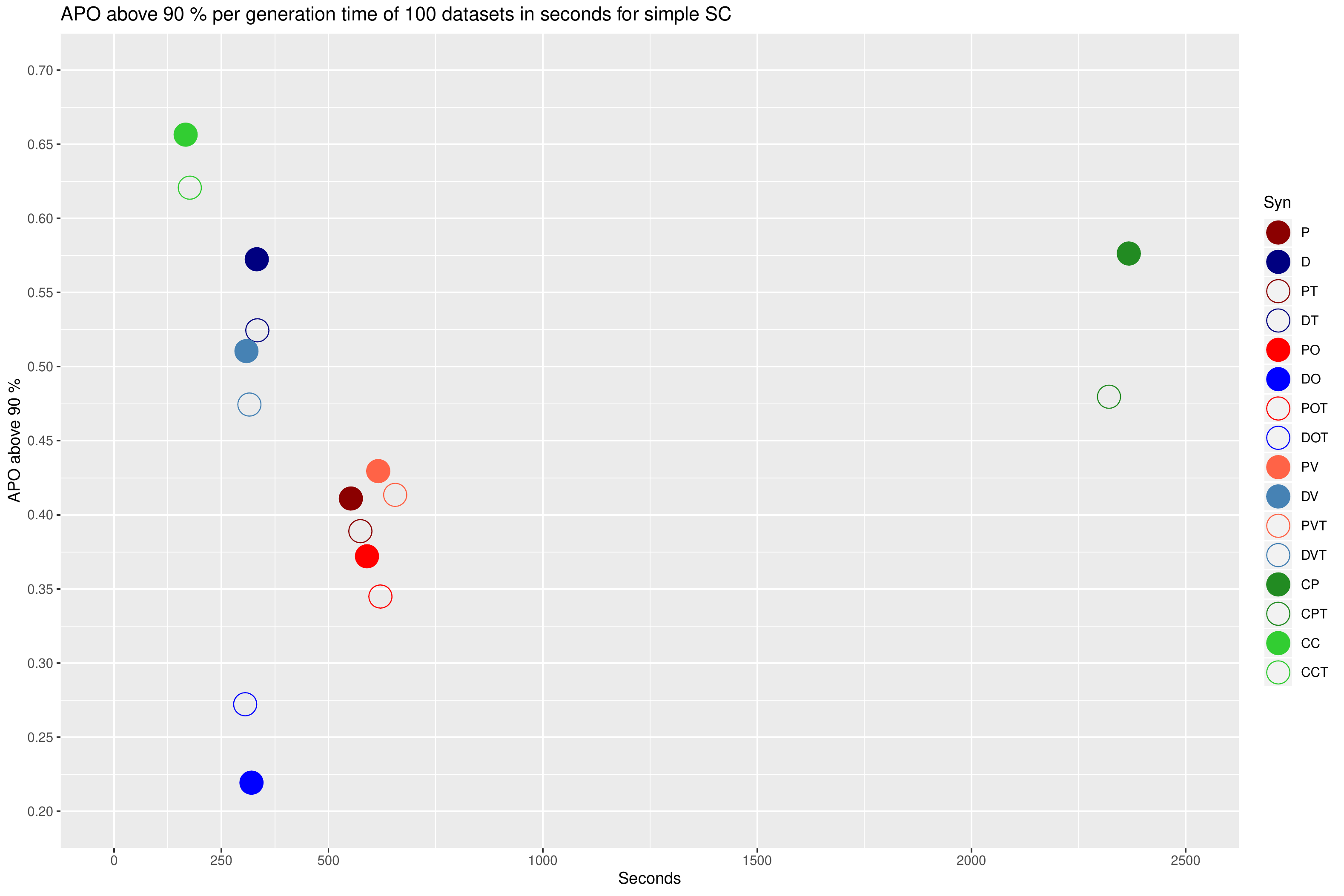}
  \caption{Polish Dataset - Simple}
  \label{fig:polish_gen_time_simp}
\end{subfigure}%
\begin{subfigure}{.33\textwidth}
  \centering
  \includegraphics[width=1\linewidth]{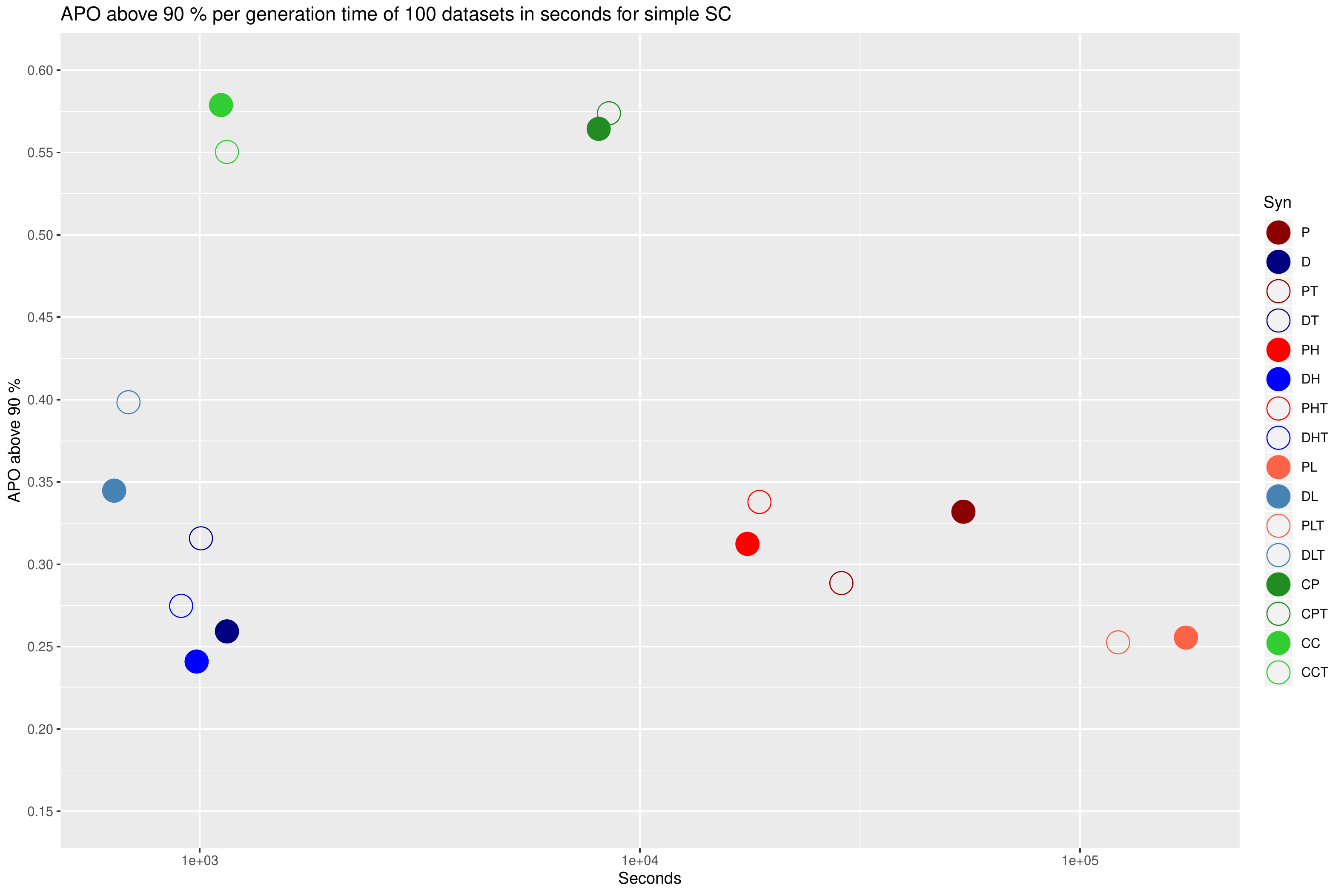}
  \caption{Adult Dataset - Simple}
  \label{fig:adult_gen_time_simp}
\end{subfigure}
\begin{subfigure}{.33\textwidth}
  \centering
  \includegraphics[width=1\linewidth]{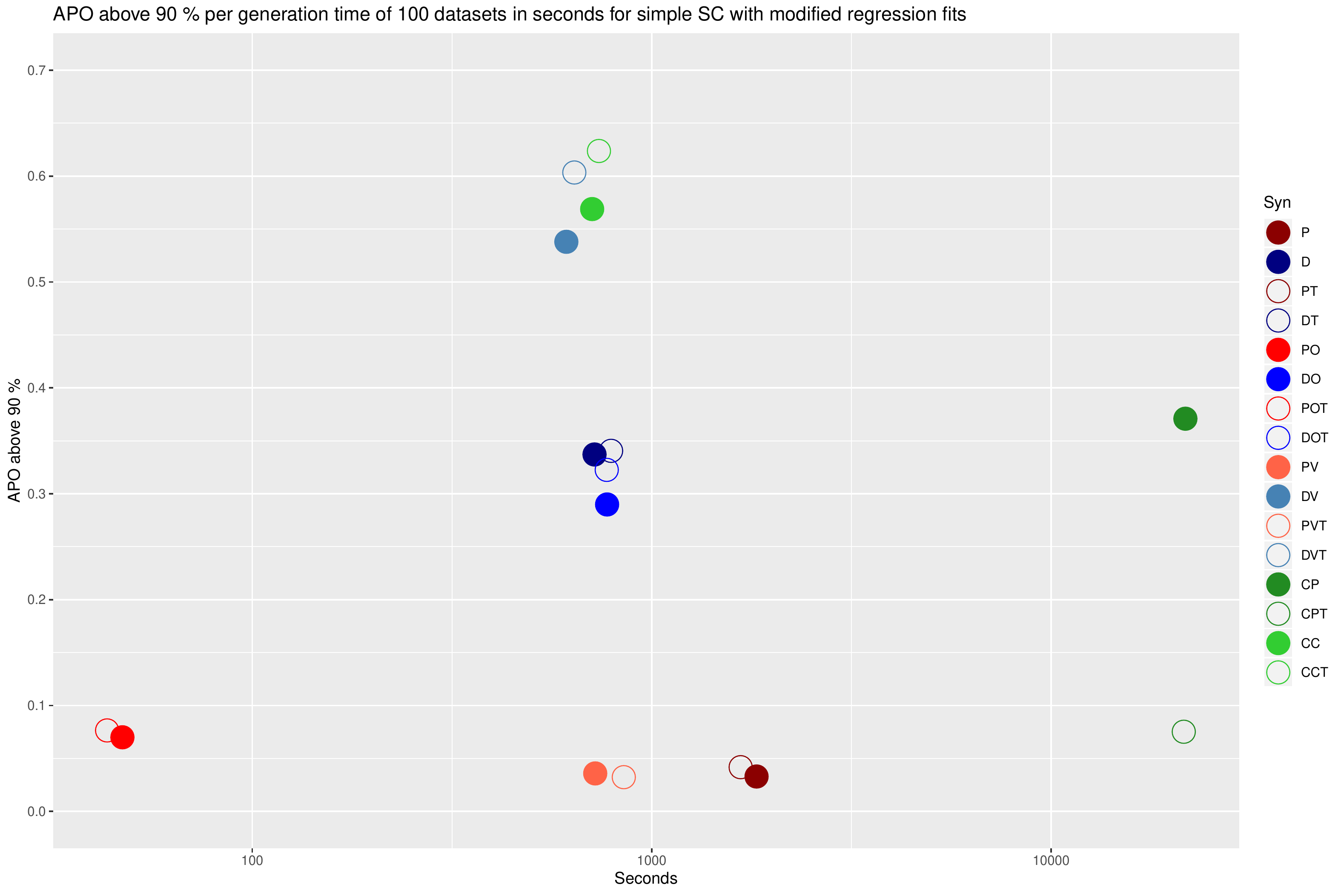}
  \caption{Avila Dataset - Simple}
  \label{fig:avila_gen_time_simp}
\end{subfigure}
}
\resizebox{\linewidth}{!}{
\begin{subfigure}{.33\textwidth}
  \centering
  \includegraphics[width=1\linewidth]{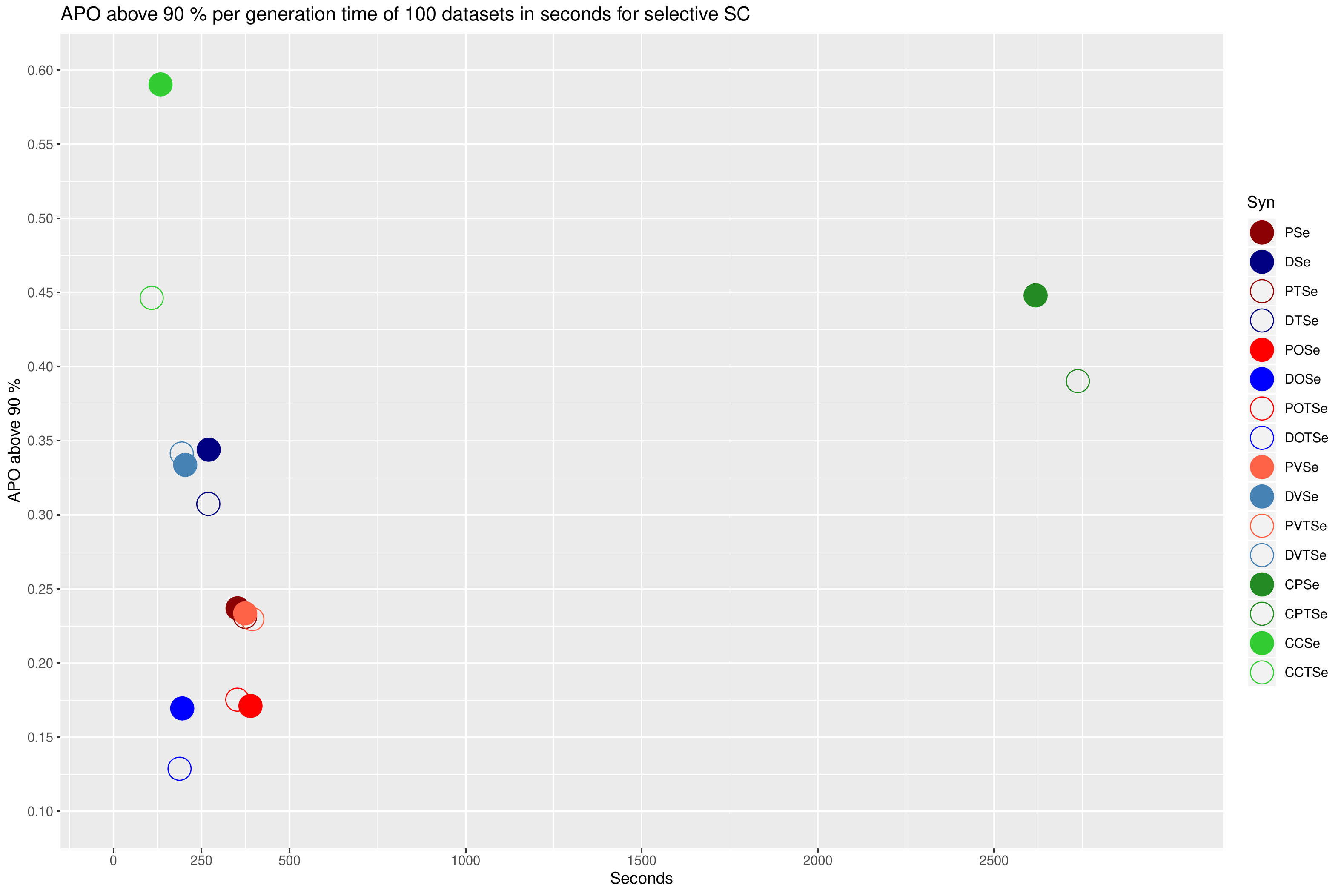}
  \caption{Polish Dataset - Selective}
  \label{fig:polish_gen_time_sel}
\end{subfigure}%
\begin{subfigure}{.33\textwidth}
  \centering
  \includegraphics[width=1\linewidth]{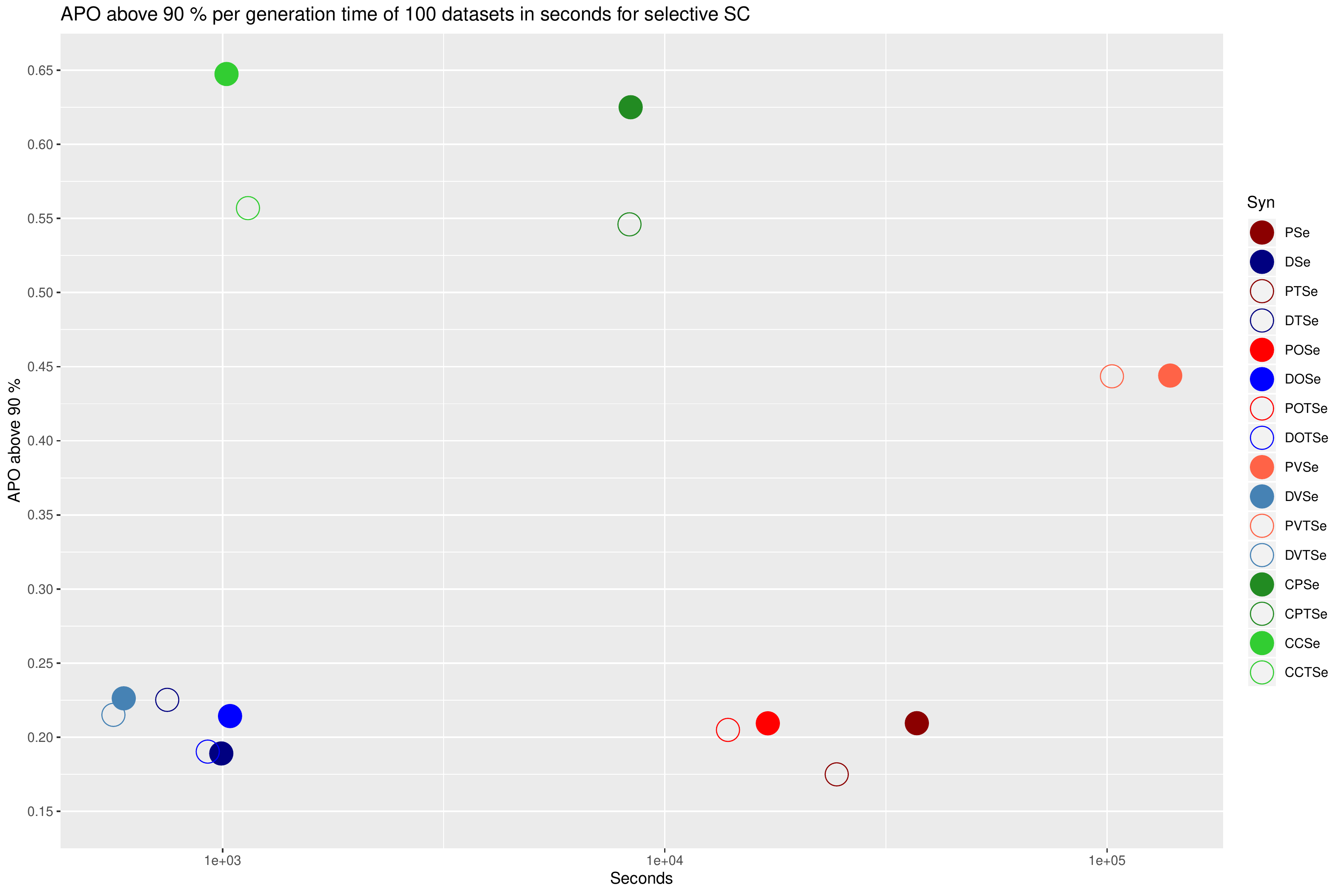}
  \caption{Adult Dataset - Selective}
  \label{fig:adult_gen_time_sel}
\end{subfigure}
\begin{subfigure}{.33\textwidth}
  \centering
  \includegraphics[width=1\linewidth]{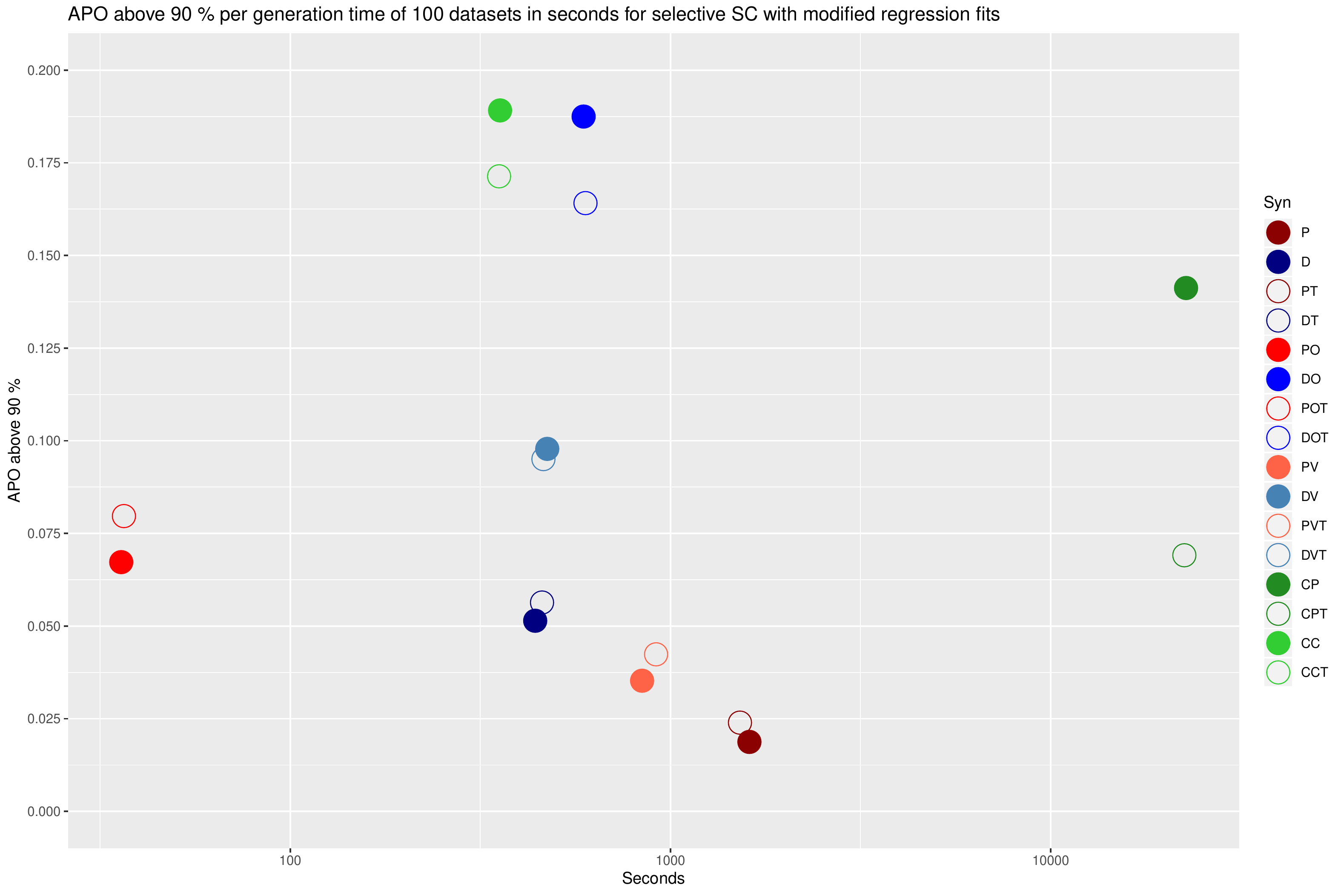}
  \caption{Avila Dataset - Selective}
  \label{fig:avila_gen_time_sel}
\end{subfigure}
}
\caption{APO above 90\% and Generation Time for 100 Datasets}
\label{fig:gen_time}
\end{figure*}

\noindent\textbf{Comparison between Simple and Selective Synthesis.}
For the comparison between Simple and Selective synthesis, we looked at the APO 90\% using both the techniques with Non-Proper synthesis and $m = 10$ (Appendix Table \ref{tab:simple_selective_short}). The results show that for majority of the synthesizers in all of the three datasets simple works better than selective. For Polish, the simple method was better regardless of the synthesizers. However, for Adult and Avila, we find that few synthesizers performed better with selective synthesis. For Adult, CP and CC with Selective performed slightly better than simple. Nonetheless, P with synthesis order largest categorical variable last (PL) achieved a significant improvement with selective. For Avila, except for P with opposite synthesis order (PO), simple performed better than selective for all the other synthesizers. Thus, in terms of accuracy simple seems to be a better choice.

\vspace{-27pt}
\begin{table*}[htbp]
\centering
\caption{APO above $90$\% Comparison between Simple and Selective with $m = 10$}
\label{tab:simple_selective_short}
\resizebox{\textwidth}{!}{%
\begin{tabular}{|l|r|cccccccccccccccc|}
\hline
\rowcolor[HTML]{EFEFEF} 
\multicolumn{1}{|c|}{\cellcolor[HTML]{EFEFEF}\textbf{Dataset}} &
  \multicolumn{1}{c|}{\cellcolor[HTML]{EFEFEF}\textbf{metric}} &
  \textbf{P} &
  \textbf{PSe} &
  \textbf{D} &
  \textbf{DSe} &
  \textbf{PO} &
  \textbf{POSe} &
  \textbf{DO} &
  \textbf{DOSe} &
  \textbf{PV} &
  \textbf{PVSe} &
  \textbf{DV} &
  \textbf{DVSe} &
  \textbf{CP} &
  \textbf{CPSe} &
  \textbf{CC} &
  \textbf{CCSe} \\ \hline
 &
  Average &
  0.38 &
  0.22 &
  0.50 &
  0.30 &
  0.33 &
  0.16 &
  0.21 &
  0.16 &
  0.39 &
  0.21 &
  0.45 &
  0.29 &
  0.49 &
  0.40 &
  0.58 &
  0.48 \\ \cline{2-18} 
\multirow{-2}{*}{Polish} &
  Ratio &
  0.78 &
  0.22 &
  0.85 &
  0.15 &
  0.83 &
  0.17 &
  0.63 &
  0.37 &
  0.76 &
  0.24 &
  0.90 &
  0.10 &
  0.65 &
  0.35 &
  0.75 &
  0.25 \\ \hline
\rowcolor[HTML]{EFEFEF} 
\multicolumn{1}{|c|}{\cellcolor[HTML]{EFEFEF}\textbf{}} &
  \multicolumn{1}{c|}{\cellcolor[HTML]{EFEFEF}\textbf{}} &
  \textbf{P} &
  \textbf{PSe} &
  \textbf{D} &
  \textbf{DSe} &
  \textbf{PH} &
  \textbf{PHSe} &
  \textbf{DH} &
  \textbf{DHSe} &
  \textbf{PL} &
  \textbf{PLSe} &
  \textbf{DL} &
  \textbf{DLSe} &
  \textbf{CP} &
  \textbf{CPSe} &
  \textbf{CC} &
  \textbf{CCSe} \\ \hline
 &
  Average &
  0.29 &
  0.20 &
  0.27 &
  0.20 &
  0.31 &
  0.20 &
  0.24 &
  0.21 &
  0.26 &
  0.39 &
  0.31 &
  0.23 &
  0.51 &
  0.55 &
  0.51 &
  0.55 \\ \cline{2-18} 
\multirow{-2}{*}{Adult} &
  Ratio &
  0.81 &
  0.19 &
  0.79 &
  0.21 &
  0.88 &
  0.12 &
  0.70 &
  0.30 &
  0.29 &
  0.71 &
  0.75 &
  0.25 &
  0.30 &
  0.70 &
  0.46 &
  0.54 \\ \hline
\rowcolor[HTML]{EFEFEF} 
\multicolumn{1}{|c|}{\cellcolor[HTML]{EFEFEF}\textbf{}} &
  \multicolumn{1}{c|}{\cellcolor[HTML]{EFEFEF}\textbf{}} &
  \textbf{P} &
  \textbf{PSe} &
  \textbf{D} &
  \textbf{DSe} &
  \textbf{PO} &
  \textbf{POSe} &
  \textbf{DO} &
  \textbf{DOSe} &
  \textbf{PV} &
  \textbf{PVSe} &
  \textbf{DV} &
  \textbf{DVSe} &
  \textbf{CP} &
  \textbf{CPSe} &
  \textbf{CC} &
  \textbf{CCSe} \\ \hline
 &
  Average &
  0.10 &
  0.08 &
  0.14 &
  0.04 &
  0.06 &
  0.08 &
  0.18 &
  0.14 &
  0.11 &
  0.10 &
  0.24 &
  0.07 &
  0.23 &
  0.10 &
  0.24 &
  0.16 \\ \cline{2-18} 
\multirow{-2}{*}{Avila} &
  Ratio &
  0.68 &
  0.32 &
  0.91 &
  0.09 &
  0.37 &
  0.63 &
  0.61 &
  0.39 &
  0.58 &
  0.42 &
  0.96 &
  0.04 &
  0.88 &
  0.12 &
  0.74 &
  0.26 \\ \hline
\end{tabular}%
}
\end{table*}

\end{document}